\documentclass{article}

\usepackage{microtype}
\usepackage{graphicx}
\usepackage{booktabs} %

\usepackage{hyperref}

\usepackage[accepted]{icml2023}

\usepackage{amsmath}
\usepackage{amssymb}
\usepackage{mathtools}
\usepackage{amsthm}

\usepackage[capitalize]{cleveref}
\Crefname{section}{Sec.}{Sections.}

\usepackage[commands,enumerate,environments,lmrscinnershape]{AVT}
\usepackage{nicefrac}
\usepackage{xspace}         %
\usepackage{dblfloatfix}    %

\usepackage{caption}
\usepackage{subcaption}

\usepackage{standalone}
\usepackage{tikz}
\usepackage{pgfplots}
\pgfplotsset{compat=1.17}
\usepgfplotslibrary{groupplots}
\usepgfplotslibrary{fillbetween}
\usetikzlibrary{fadings}
\definecolor{purple148103189}{RGB}{148,103,189}
\definecolor{forestgreen4416044}{RGB}{44,160,44}
\definecolor{darkorange25512714}{RGB}{255,127,14}

\usepackage{annotate-equations}

\theoremstyle{plain}
\newtheorem{theorem}{Theorem}[section]
\newtheorem{proposition}[theorem]{Proposition}
\newtheorem{lemma}[theorem]{Lemma}

\theoremstyle{definition}
\newtheorem{definition}[theorem]{Definition}

\theoremstyle{remark}

\usepackage[textsize=tiny]{todonotes}

\icmltitlerunning{Neural Diffusion Processes}

\begin{document}

\twocolumn[
\icmltitle{Neural Diffusion Processes}

\begin{icmlauthorlist}
\icmlauthor{Vincent Dutordoir}{cambridge,secondmind}
\icmlauthor{Alan Saul}{secondmind}
\icmlauthor{Zoubin Ghahramani}{cambridge,brain}
\icmlauthor{Fergus Simpson}{secondmind}
\end{icmlauthorlist}

\icmlaffiliation{cambridge}{Department of Engineering, University of Cambridge, Cambridge, UK}
\icmlaffiliation{secondmind}{Secondmind, Cambridge, UK}
\icmlaffiliation{brain}{Google DeepMind}

\icmlcorrespondingauthor{Vincent Dutordoir}{vd309@cam.ac.uk}

\icmlkeywords{score-based generative modelling, diffusion model, neural processes, Gaussian processes}

\vskip 0.3in
]

\printAffiliationsAndNotice{}

\begin{abstract}
Neural network approaches for meta-learning distributions over functions have desirable properties such as increased flexibility and a reduced complexity of inference. Building on the successes of denoising diffusion models for generative modelling, we propose Neural Diffusion Processes (NDPs), a novel approach that learns to sample from a rich distribution over functions through its finite marginals. By introducing a custom attention block we are able to incorporate properties of stochastic processes, such as exchangeability, directly into the NDP's architecture. We empirically show that NDPs can capture functional distributions close to the true Bayesian posterior, demonstrating that they can successfully emulate the behaviour of Gaussian processes and surpass the performance of neural processes. NDPs enable a variety of downstream tasks, including regression, implicit hyperparameter marginalisation, non-Gaussian posterior prediction and global optimisation.
\end{abstract}

\section{Introduction}
Gaussian processes (GPs) offer a powerful framework for defining distributions over functions \citep{rasmussen06}. It is an appealing framework because
Bayes rule allows one to reason consistently about the predictive distribution, allowing the model to be data efficient. However, for many problems, GPs are not an appropriate prior. Consider, for example, a function that has a single discontinuity at some unknown location. This is one classic example of a distribution of functions that cannot be expressed in terms of a GP \citep{Neal1998Regression}.

One popular approach to these problems is to abandon GPs, in favour of Neural Network-based generative models. Successful methods include the meta-learning approaches of Neural Processes (NPs) \citep{garnelo2018neural,garnelo2018conditional}, and VAE-based models \citep{mishra2020pi,fortuin2020gp}. By leveraging a large number of small datasets during training, they are able to transfer knowledge across datasets at prediction time. Using Neural Networks (NNs) is appealing since most of the computational effort is expended during the training process, while the task of prediction becomes more straightforward. A further major advantage of a NN-based approach is that they are not restricted by Gaussianity.

In this work, we strive to enhance the capabilities of NN-based generative models for meta-learning function by extending probabilistic denoising diffusion models \citep{sohl2015deep,song2020improved,ho2020denoising}. Diffusion models have demonstrated superior performance over existing methods in generating images \citep{nichol2021glide, ramesh2022hierarchical}, molecular structures \citep{xu2022geodiff, hoogeboom22Equivariant}, point clouds \citep{luo2021diffusion}, and audio signal data \cite{kong2020diffwave}. The central challenge we address is the Bayesian inference of functions, a fundamentally different task not previously approached by diffusion models. Although we acknowledge that several recent works have emerged since our initial preprint, providing complementary insights on extending diffusion models to function spaces \citep{phillips2022Spectral,kerrigan2022Diffusion,lim2023score,bond2023infty,pidstrigach2023InfiniteDimensional,franzese2023continuous}.

\paragraph{Contributions} We propose a novel generative model, the Neural Diffusion Process (NDP), which defines a probabilistic model over functions via their finite marginals. NDPs generalise diffusion models to stochastic processes by allowing the indexing of the function's marginals onto which the model diffuses. We take particular care to enforce properties of stochastic processes, including exchangeability, facilitating the training process. These properties are enforced using a novel \emph{bi-dimensional} attention block, which guarantees equivariance over the ordering of the input dimensionality and the sequence (i.e., datapoints). From the experiments, we conclude: firstly, NDPs are an improvement over existing NN-based generative models for functions such as Neural Processes (NPs). Secondly, NDPs are an attractive alternative to GPs for specifying appropriate (i.e., non-Gaussian) priors over functions. Finally, we present a novel global optimisation method using NDPs.

\section{Background}
\label{sec:backgroud}

The aim of this section is to provide an overview of the key concepts used throughout the manuscript.

\begin{figure*}
\centering
\begin{subfigure}{.32\textwidth}
\centering
\includegraphics[width=\textwidth]{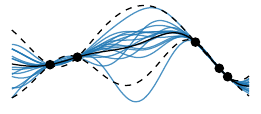}
\caption{GP Regression}
\label{fig:one-d-regression-gp-single}
\end{subfigure}
\begin{subfigure}{.32\textwidth}
\centering
\includegraphics[width=\textwidth]{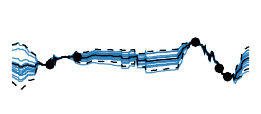}
\caption{Attentive Latent NP}
\label{fig:one-d-regression-np-single}
\end{subfigure}
\begin{subfigure}{.32\textwidth}
\centering
\includegraphics[width=\textwidth]{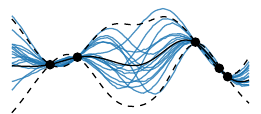}
\caption{Neural Diffusion Process (ours)}
\label{fig:one-d-regression-ndp-single}
\end{subfigure}
\caption{Posterior samples conditioned on a context dataset (black dots) for different probabilistic models.} %
\label{fig:one-d-regression-single}
\end{figure*}

\subsection{Gaussian Processes}
\label{sec:backgroud:gaussian-processes}

A Gaussian Process (GP) $f: \R^D \rightarrow \R$ is a stochastic process such that, for any finite collection of points $x_1, ..., x_n \in \R^D$ the random vector $(f_1,\dots, f_n)$ with $f_i = f(x_i)$, follows a multivariate normal distribution \citep{rasmussen06}. For a dataset $\c{D} = \{(x_i, y_i)\}_{i=1}^n$, where values are corrupted by Gaussian noise $y_i = f(x_i) + \eta$, GPs offer exact inference of the posterior $p(\v{y}^* \given \c{D})$. This leads to data-efficient learning and accurate uncertainty estimations as illustrated in \cref{fig:one-d-regression-gp-single}.

GPs are stochastic processes (SPs) which satisfy the Kolmogorov Extension Theorem (KET). KET states that all finite-dimensional marginal distributions $p$ are consistent with each other under permutation (exchangeability) and marginalisation. Let $n\in \bb{N}$ and $\pi$ be a permutation of $\{1, \ldots, n\}$, then the following holds for the GP's joint:
\begin{align}
    &p(f_1, \ldots, f_n) = p(f_{\pi(1)}, \ldots, f_{\pi(n)}),\text{ and} \label{eq:exchangeability}\\
    &p(f_1) = \int p(f_1, f_2,\ldots,f_n) \d f_2 \ldots \d f_n. \label{eq:marginal-consistency}
\end{align}
Despite these favourable properties, GPs are plagued by several limitations. 
Firstly, encoding prior assumptions through analytical covariance functions can be extremely difficult, especially in higher dimensions \citep{wilson2016deep,simpson2021kernel,liu2020task}. 
Secondly, by definition, GPs assume a multivariate \emph{Gaussian} distribution for each finite collect of predictions ---limiting the set of functions it can model \citep{Neal1998Regression}. %

\subsection{Neural Processes and Meta-Learning Functions}
\label{sec:backgroud:neural-processes}

Neural Process (NP) models were introduced as a flexible alternative to GPs, utilising an encoder-decoder architecture to meta-learn dataset distributions in an amortised fashion \citep{garnelo2018neural}. While being beneficial in many respects, they nonetheless present several limitations: Conditional NPs assume independence amongst all function values in the predictive posterior, leading to a lack of correlated function samples at test time \citep{garnelo2018conditional}. Latent NPs tackle this but have non-tractable likelihoods, leading to crude approximations during inference and performance restrictions. Attentive NPs improve empirical performance using attention mechanisms, yet frequently produce jittery samples due to shifting attention patterns \citep{kim2019attentive}. Furthermore, NPs fail to ensure consistency in the context sets as their conditional distributions do not relate through Bayes' rule \citep{kim2019attentive}. We refer to \cref{sec:app:nps} for a primer on these methods.

Several models have sought to address these limitations. ConvCNPs enforce stationary into the stochastic process by decoding functional embeddings using convolutional NNs \citep{Gordon2019Convultional}. Gaussian NPs model predictive correlations (i.e. covariances), which provides universal approximation guarantees but struggle with computational scalability \citep{bruinsma2021gaussian}. Recently, Autoregressive NPs were introduced which exhibit promising performance on a range of problems but are hampered by simple distributions, typically Gaussian, for variables early in the auto-regressive generation \citep{bruinsma2023autoregressive}.

\subsection{Probabilistic Denoising Diffusion Models}
\label{sec:backgroud:diffusion-models}

Diffusion models depend on two procedures: a \emph{forward} and a \emph{reverse} process. The forward process consists of a Markov chain, which incrementally adds random noise to the data. The reverse process is tasked with inverting this chain. The \emph{forward process} starts from the data distribution $q(\v{s}_0)$ and iteratively corrupts samples by adding small amounts of noise, which leads to a fixed Markov chain $q(\v{s}_{0:T}) = q(\v{s}_0)\prod_{t=1}^T q(\v{s}_{t} \given \v{s}_{t-1})$ for a total of $T$ steps. The corrupting distribution is Gaussian $q(\v{s}_{t} \given \v{s}_{t-1}) = \c{N}(\v{s}_{t}; \sqrt{1-\beta_t} \v{s}_{t-1}, \beta_t \m{I})$. The magnitude of the noise in each step is controlled by a pre-specified variance schedule $\{\beta_t \in (0, 1)\}_{t=1}^T$. Note that there is no learning involved in the forward process, it simply creates a sequence of random variables $\{\v{s}_t\}_{t=0}^T$ which progressively look more like white noise 
$q(\v{s}_T) \approx \c{N}(\v{0}, \m{I})$.

While the forward process is Markovian, the true reverse probability $q(\v{s}_{t-1} \given \v{s}_{t})$ requires the entire sequence. Therefore, the \emph{reverse process} learns to approximate these conditional probabilities in order to carry out the reverse diffusion process. The approximation relies on the key observation that the reverse conditional probability is tractable when conditioned on the initial state $\v{s}_0$: $q(\v{s}_{t-1} \given \v{s}_{0}, \v{s}_{t}) = \c{N}(\v{s}_{t-1}; \tilde{\v{\mu}}(\v{s}_{0}, \v{s}_{t}), \tilde{\beta}_t \m{I})$. As a result, the reverse process can be traversed by estimating the initial state $\v{s}_0$ from $\v{s}_t$ and $t$ using a NN, which can then be passed to $\tilde{\v{\mu}}$. In this work, we follow \citet{ho2020denoising} who directly parameterise $\tilde{\v{\mu}}$ as $\v{\mu}_\theta(\v{s}_{t}, t) = \frac{1}{\sqrt{\alpha_t}} \big(\v{s}_t - \frac{\beta_t}{\sqrt{1 - \bar{\alpha}_t}} \v{\epsilon}_\theta(\v{s}_t, t)\big)$ with $\alpha_t = 1 - \beta_t$ and $\bar{\alpha}_t = \prod_{j=1}^t \alpha_j$. In what follows, we refer to $\v{\epsilon}_\theta$ as the noise model. %
The parameters $\theta$ of the noise model are optimised by minimising the objective $\E_{t, \v{s}_0, \v{\varepsilon}}[\v{\varepsilon} - \v{\epsilon}_\theta(\v{s}_t, t)]$, where the expectation is taken over time $t \~ \c{U}(\{1,2,\ldots, T\})$, data $\v{s}_0 \~ q(\v{s}_0)$, and $\v{s}_t = \sqrt{\bar{\alpha}_t} \v{s}_0 + \sqrt{1-\bar{\alpha}_t} \v{\varepsilon}$ with $\v{\varepsilon} \~ \c{N}(\v{0}, \m{I})$. %
For a trained network, generating samples from $q(\v{s}_0)$ is done by running the reverse process starting from $\v{s}_T \~ \c{N}(\v{0}, \m{I})$. %

\section{Neural Diffusion Processes}

In this section, we introduce Neural Diffusion Processes (NDPs), which define a probabilistic model over functions via their finite marginals. We focus our attention on the difference between NDPs and traditional diffusion models used for gridded data, such as images or audio. %

\subsection{Data, Forward and Reverse Process}

NDPs generalise diffusion models to stochastic processes by allowing indexing of the random variables which are being diffused. This set of random variables corresponds to the function's marginals, which represent a more flexible type of random variable compared to, say, an image. In an image, pixel values are organised on a predefined grid (height $\times$ width) with an implicit order. Contrarily, function values lack ordering and do not reside on a predetermined grid. Consequently, samples extracted from a NDP should be evaluable throughout their input domain.

\paragraph{Data} NDPs require many datasets that consist of both, function inputs $\v{x}$ and corresponding function values $\v{y} = f(\v{x})$. The functions $f: \R^D \rightarrow \R$ may be synthetically created (e.g., step functions, \cref{sec:experiment:step}), drawn from a Gaussian process (\cref{sec:experiment:regression}), or correspond to the pixel maps in images (\cref{sec:experiment:image-regression}). Crucially, in this meta-learning approach, the NDP needs numerous examples drawn from a distribution over $f$ to learn an empirical covariance from the data, $\{(\v{x}^i \in \R^{N \times D}, \v{y}^i \in \R^N)\}_{i=1}^M$ where $\v{y}^i = f^i(\v{x}^i)$ assuming we evaluate the function at $N$ random location in its domain. This setup is akin to NPs, the key difference being that NDPs do not necessitate the separation of the dataset into a context and target set during training.

\paragraph{Forward process} Let $\v{x}_0 = \v{x}$ and $\v{y}_0 = \v{y}$ then NDPs gradually add noise following
$$
\label{eq:forward}
    q\big(\colvec{\v{x}_{t}}{\v{y}_{t}}\!\given\!\colvec{\v{x}_{t-1}}{\v{y}_{t-1}} \big) \!\!= \c{N}\big(\v{y}_{t}; \sqrt{1-\beta_t} \v{y}_{t-1}, \beta_t\big).
$$
This corresponds to adding Gaussian noise to the function values $\v{y}_t$ while keeping the input locations $\v{x}_t$ fixed across time. At $t=T$ the function values $\v{y}_t$ should be indistinguishable to samples from $\c{N}(\v{0}, \m{I})$, while $\v{x}_T = \v{x}_{T-1} = \ldots = \v{x}_0$. In \cref{sec:experiment:joint}, we discuss a generalisation of this scheme where we perturb both the function inputs and outputs to obtain a diffusion model that can sample the joint $p(\v{x}_0, \v{y}_0)$. For now, we focus on the conditional $p(\v{y}_0 \given \v{x}_0)$ as this is the distribution of interest in supervised learning.

\paragraph{Backward kernel}
NDPs parameterise the backward Markov kernel using a NN that learns to de-noise the corrupted function values $\v{y}_t$. In contrast, the input locations $\v{x}_t$ have not been corrupted in the forward process which makes reversing their chain trivial. This leads to a parameterised backward kernel $p_\theta$ of the form
\begin{equation}
\label{eq:backward}
p_\theta\big(\colvec{\v{x}_{t-1}}{\v{y}_{t-1}} \given \colvec{\v{x}_{t}}{\v{y}_{t}}\big) =
\c{N}\Big(\v{y}_{t-1}; \v{\mu}_\theta(\v{x}_t, \v{y}_t, t), \tilde{\beta}_t \Big),
\end{equation}
where the mean is parameterised through the noise $\v{\epsilon}_\theta$ as
\begin{equation}
\label{eq:parameterised-mean}
    \v{\mu}_\theta(\v{x}_t, \v{y}_t, t) = \frac{1}{\sqrt{\alpha_t}} \big(\v{y}_t - \frac{\beta_t}{\sqrt{1 - \bar{\alpha}_t}} \v{\epsilon}_\theta(\v{x}_t, \v{y}_t, t)\big).
\end{equation}
The noise model NN $\v{\epsilon}_\theta: \R^{N \x D} \x \R^N \x \R \rightarrow \R^N$ has as inputs the function inputs $\v{x}_t$, the corrupted function values $\v{y}_t$, and time $t$. The network is tasked with predicting the noise that was added to $\v{y}_0$ to obtain $\v{y}_t$. The design of the network is specific to the task of modelling functions and as such differs from current approaches. In the next section we discuss its particular architecture, but for now we want to stress that it is critical for the noise model to have access to the input locations $\v{x}_t$ to make such predictions.

\paragraph{Objective} Following \citet{ho2020denoising}, but substituting the NDP's forward and backward transition densities leads to the objective
\begin{equation}
\label{eq:objective}
\c{L}_\theta = \E_{t, \v{x}_0, \v{y}_0, \v{\varepsilon}}\Big[\|\v{\varepsilon} - \v{\epsilon}_\theta\big(\v{x}_0, \v{y}_t, t\big)\|^2\Big],
\end{equation}
with $\v{y}_t = \sqrt{\bar{\alpha}_t} \v{y}_0 + \sqrt{1-\bar{\alpha}_t} \v{\varepsilon}$ and where we made use of the fact that $\v{x}_t$ equals $\v{x}_0$ for all timesteps $t$. Full derivation of the objective is given in~\cref{sec:app:derivation-lower-bound}.

\begin{figure*}[t]
    \centering
    \includegraphics[width=\textwidth]{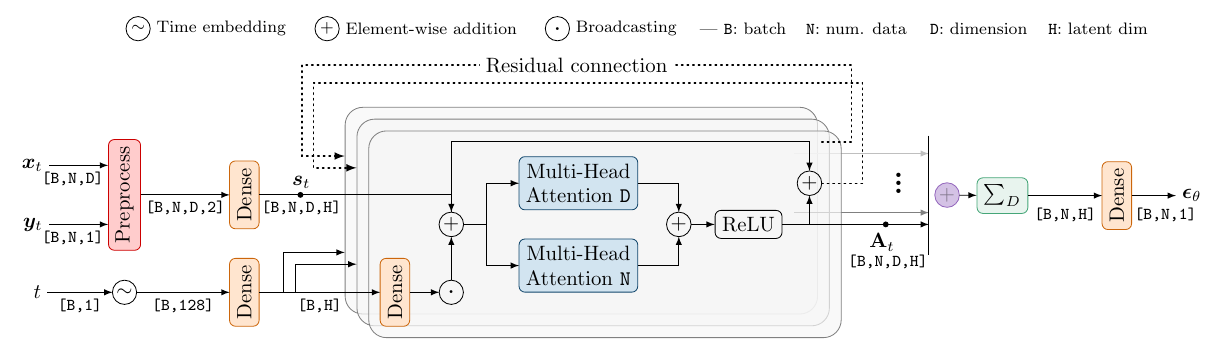}
    \caption{Architecture of the noise prediction model, utilised at each step within the Neural Diffusion Process. The greyed box represents the bi-dimensional attention block, as discussed in Section \ref{sec:bidim}.}
    \label{fig:architecture}
\end{figure*}

\subsection{Prior and Conditional Sampling}
\label{sec:prior-conditional-sampling}

\paragraph{Prior} Using a trained noise model $\v{\epsilon}_\theta$, one can obtain prior function draws from the NDP at a specific set of input locations $\v{x}_0$ by simulating the reverse process. That is, starting from $\v{y}_T \~ \c{N}(\v{0}, \m{I})$ and iteratively using the backward kernel $p_\theta$ from \cref{eq:backward} for time $t=T,\ldots,1$. This procedure leads to the samples from the prior illustrated in the left panes of \cref{fig:kernel-marginalisation,fig:step}.

\paragraph{Conditional} NDPs draw samples from the conditional distribution $p(\v{y}^*_0, \given \v{x}^*_0, \c{D})$, where $\c{D} = (\v{x}_0^c \in \R^{M \x D}, \v{y}_0^c \in \R^M)$ is the \emph{context} dataset, using a slight adaptation of {\scshape RePaint} algorithm \citep{lugmayr2022repaint}.

The conditional sampling scheme, given as pseudocode in \cref{sec:app:code-conditional-sampling}, works as follows. Start by sampling $\v{y}_T^* \~ \c{N}(\v{0}, \m{I})$. Then, for each time $t$, sample a noisy version of the context $\v{y}_t^c$ using the forward process 
\begin{equation}
\label{eq:conditional-sampling}
\v{y}_t^c \~ \c{N}\Big(
\sqrt{\bar{\alpha}_t} \v{y}_0^c, (1-\bar{\alpha}_t)\,\m{I}
\Big).
\end{equation}
Continue by collecting the union of the noisy context and target set in $\v{y}_t = \{\v{y}_t^*, \v{y}_t^c\}$. Similarly, collect the inputs as $\v{x}_0 = \{\v{x}_0^*, \v{x}_0^c\}$. Finish the step by sampling from the backward kernel $p_\theta$ using the collected inputs $\v{x}_0$ and function values $\v{y}_t$
\begin{equation}
\label{eq:conditional-sampling-2}
\v{y}_{t-1} \~ \c{N}\Big(  \frac{1}{\sqrt{\alpha_t}}\big(\v{y}_{t} - \frac{\beta_t}{\sqrt{1-\bar{\alpha}_t}}\epsilon_\theta({\v{x}}_0, {\v{y}}_t, t)\big), \tilde{\beta}_t \m{I}
\Big).
\end{equation}
Simulating this scheme from $t=T,\ldots,1$ ensures that for each backward step we leverage the context dataset. Consistent with the findings of \citet{lugmayr2022repaint}, we found that in practice the sample quality improves by repeating \cref{eq:conditional-sampling,eq:conditional-sampling-2} multiple times (e.g., 5) per time step. We show conditional samples in \cref{fig:one-d-regression-ndp-single,fig:kernel-marginalisation,fig:step} using our proposed algorithm.

\section{Noise Model Architecture}
\label{sec:architecture}

NDPs implement the noise model $\v{\epsilon}_\theta$ as a NN. Here, we review its architecture and key components. In principle, \emph{any} NN could be used. However, if we wish for NDPs to mimic stochastic processes (SPs), the noise model must learn to generate a prior distribution over functions. We expect such a prior to possess several key symmetries and properties, which will heavily influence our choice of architecture. %

\subsection{Input Size Agnosticity}
Before addressing the NN invariances and equivariances, we focus our attention to a key property of the network: the NDP network is agnostic to dataset size $N$ and dimension $D$. That is, the weights of the network do \emph{not} depend on the size of the inputs (i.e. $N$ nor $D$). This has as important practical consequences that it is not required to train different NDPs for datasets with different size or dimensionality. This makes it possible to train only a single model that handles downstream tasks with different $N$ or $D$ values.  To achieve this functionality, NDPs start by reshaping the inputs $(\v{x}_t \in \R^{N \times D}, \v{y}_t \in \R^N)$ to $\R^{N \x D \x 2}$ by replicating the $\v{y}_t$ outputs $D$ times before concatenating them with $\v{x}_t$.

\subsection{Bi-Dimensional Attention Block}
\label{sec:bidim}
A key property of stochastic processes is an \emph{equivariance} to the ordering of inputs. As such, shuffling the order of data points in the context dataset $\c{D}$ or the order at which we make predictions should not affect the probability of the data (i.e.,~the data is exchangeable). Secondly, we also expect an \emph{invariance} in the ordering of the input dimensions. Consider, for example, a dataset consisting of two features, the weight and height of people. We would not expect the posterior function to be different if we would swap the order of the columns in the training data. This is an important invariance encoded in many GP kernels (e.g., Mat\'ern, RBF) but often overlooked in the neural net literature. %

We accommodate for both desiderata using our proposed \emph{bi-dimensional attention block}. We denote this block by $\m{A}_t: \R^{N\times D\times H} \rightarrow \R^{N \times D \times H}$ as it acts on the preprocessed inputs $(\v{x}_t, \v{y}_t)$. At its core, the block consists of two multi-head self-attention (MHSA) layers \citep{vaswani2017attention} (see \cref{fig:architecture}). The MHSA layers act on different axes of the input: one attends to the input dimension axis $d$, while the other attends across the dataset sequence axis $n$. The outputs of the two are subsequently summed and passed through a non-linearity. This process is repeated multiple times by feeding the output back into the next bi-dimensional attention block using residual connections. Concretely, the $\ell^{\textrm{th}}$ block is defined as
\begin{equation*}
    \m{A}^\ell_t(\v{s}_t^{\ell-1}) = \m{A}_{t}^{\ell-1} + \sigma\left(\textrm{MHSA}_d(\v{s}^{\ell-1}_t) + \textrm{MHSA}_n(\v{s}^{\ell-1}_t)\right),
\end{equation*}
for $\ell=\{1, \ldots, L\}$ with $\sigma$ the ReLU function, $\v{s}_t^{0} = \v{s}_t$ (i.e., the output of the preprocessing step) and $\m{A}_t^0=\v{0}$. \citet{kossen2021self} introduced a similar block which operates sequentially across datapoints and attributes, whereas ours acts in parallel.

To summarise, the bi-dimensional attention block is \emph{equivariant} to the order of the dimensions and dataset sequence, as formalised by
\begin{proposition}
\label{prop:one}
Let ${\Pi}_{N}$ and ${\Pi_D}$ be the set of all permutations of indices $\{1, \ldots, N\}$ and $\{1, \ldots, D\}$, respectively. Let $\v{s} \in \R^{N \times D \times H}$ and $(\pi_n \circ \v{s}) \in \R^{N \times D \times H}$ denote a tensor where the ordering of indices in the \emph{first} dimension are given by $\pi_n \in \Pi_N$.
Similarly, let $(\pi_d \circ \v{s})$ denote a tensor where the ordering of indices in the \emph{second} dimension are given by $\pi_d \in \Pi_D$. Then, $\forall \pi_n,\pi_d \in \Pi_N \x \Pi_D$:
\begin{equation}
    \pi_d \circ \pi_n \circ \m{A}_t(\v{s}) = \m{A}_t\left(\pi_d \circ \pi_n \circ \v{s} \right).
\end{equation}
\end{proposition}
\begin{proof}
Follows from the invariance of MHSA and the commutativity of $\pi_d$ and $\pi_n$ as detailed in \cref{sec:app:proof-bi-dimensional-attention-block}.
\end{proof}

The final noise model output $\v{\epsilon}_{\theta}$ is obtained by summing the output of the different bi-dimensional attention layers (see purple `$+$' in \cref{fig:architecture}). This is followed by a sum over the input dimension axis (green block). These final operations introduce an invariance over input dimensionality, while preserving the equivariance over the dataset ordering. We summarise the noise model's invariance and equivariance properties in the following proposition
\begin{proposition}
\label{prop:exchangeability}
Let $\pi_n$ and $\pi_d$ be defined as in \Cref{prop:one}, then $\v{\epsilon}_\theta$ satisfies
\begin{equation}
\pi_n \circ \v{\epsilon}_{\theta}(\v{x}_t, \v{y}_t, t) = \v{\epsilon}_{\theta}(\pi_n \circ \pi_d \circ \v{x}_t, \pi_n \circ \v{y}_t, t).
\end{equation}
\end{proposition}
\begin{proof}
The claim follows from \Cref{prop:one} and \citet{zaheer2017deep} as shown in \cref{sec:app:neural-network-properties-final}.
\end{proof}
Crucially, by directly encoding these properties into the noise model, the NDP produces a set of random variables $\{y_t^1, \ldots, y_t^n\}$ at each time step $t$ that are exchangeable as defined in \cref{eq:exchangeability}. %

\section{On Meta-Learning Consistency}

\begin{figure*}[!t]
\centering
\includegraphics[width=\textwidth]{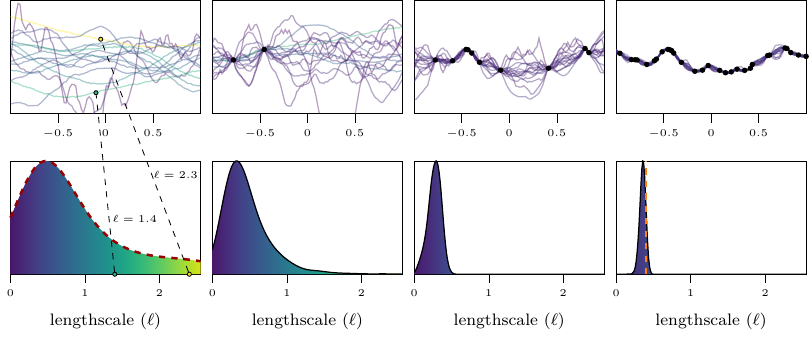}
\caption{{Hyperparameter marginalisation}: Samples from the NDP, conditioned on an increasing number of data points (black dots), are illustrated in the top row. A sample is coloured according to its most likely lengthscale. The bottom row shows a histogram of likely lengthscales from the produced samples. As more data points are provided, the distribution of likely lengthscales converges from the prior over lengthscales to the lengthscale that was used to produce the data ($\ell = 0.3$).}
\label{fig:kernel-marginalisation}
\end{figure*}

Neural Diffusion Processes (NDPs) are generative models that define a probabilistic model over functions via their finite marginals. As prescribed by the Kolmogorov extension theorem, these finite marginals originate from a stochastic process if they satisfy the conditions of exchangeability and consistency. \Cref{prop:exchangeability} has established that the exchangeability condition can be achieved by parametrising the noise model using the proposed bi-dimensional attention model, rendering the network permutation-equivariant. Consequently, we direct our focus to the more intricate topic of marginal consistency.

We first note that when the true noise $\v{\varepsilon}$ is accessible, the generative model described by the reverse process exhibits marginal consistency across all finite marginals, thereby corresponding to samples from a stochastic process. However, in NDPs the noise model is approximated by a neural network $\v{\epsilon}_\theta \approx \v{\varepsilon}$. This approximation leads to NDPs forfeiting consistency within the generative process $p_\theta$.

Even though NDPs cannot guarantee consistency as per \cref{eq:marginal-consistency}, they do embed a particular form of it. Consider ${\boldsymbol{y}} = \boldsymbol{y}^* \cup \boldsymbol{y}^c = {\boldsymbol{y}^*}' \cup {\boldsymbol{y}^{c}}'$ where, $\boldsymbol{y}^c$ and ${\boldsymbol{y}^c}'$ denote two sets of contexts and $\boldsymbol{y}^*$ and ${\boldsymbol{y}^*}'$ represent the targets. When the union of these sets are identical, NDPs guarantee that the conditionals $p(\boldsymbol{y}^*|\boldsymbol{y}^c)$ and $p({\boldsymbol{y}^*}' | {\boldsymbol{y}^c}')$ are consistent among each other and the joint $p(\boldsymbol{y})$. Consequently, NDPs can generate $2^{|\boldsymbol{y}|}$ consistent marginals from this joint. 

While this is arguably a restricted form of consistency, it is a property that (A)NPs do not possess \citep{kim2019attentive}.  As detailed in \cref{sec:app:nps}, when it comes to conditioning, NDPs are unlike (A)NPs. In (A)NPs the predictive distribution over the targets is learnt in an amortised fashion by a neural network, using the context as inputs. NDPs, on the contrary, draw conditional samples using the joint distribution over contexts and targets, as described in \cref{sec:prior-conditional-sampling}.

Continuing the example, in cases where the unions of the target and context sets in NDPs are not equal, i.e. $\boldsymbol{y}^* \cup \boldsymbol{y}^c \neq {\boldsymbol{y}^*}' \cup {\boldsymbol{y}^{c}}'$, NDPs cannot ensure consistency and must resort to meta-learning for an approximation. It has been posited by \citet{foong2020MetaLearning}, that it is in practice often more beneficial to refrain from enforcing consistency into the network as this limits the choices of network architectures, and can lead to constraints on the learning performance. This view is confirmed by our experiments, which show that NDP's predictive performance closely matches that of optimal models, despite their lack of KET consistency.

\section{Experimental Evaluation}
\label{sec:experiments}

In this section we aim to address the following two questions: Firstly, what additional advantages do NDPs offer beyond GPs? Secondly, how do NDPs stand in relation to NPs concerning performance and applicability? Furthermore, we introduce an innovative method for global optimisation of black-box functions in \cref{sec:global-optimisation}, which is predicated on modelling the joint distribution $p(x, y)$. All experiments (except \cref{sec:global-optimisation}) share the same model architecture illustrated in~\cref{fig:architecture}. Comprehensive details regarding the specific model configurations and training times can be found in~\cref{sec:app:experimental-setup}.

\subsection{Emulating Gaussian Processes}
\label{sec:ndp-vs-gp}

\subsubsection{Hyperparameter marginalisation}

Conventionally, GP models optimise a point estimate of the hyperparameters. However, it is well known that marginalising over the hyperparameters can lead to significant performance improvements, albeit at an extra computational cost \cite{lalchand2020approximate, simpson2021marginalised}. A key capability of the NDP is its ability to produce realistic conditional samples across the full gamut of data on which it was trained. It therefore in effect marginalises over the different hyperparameters it was exposed to, or even different kernels. To demonstrate this, we train a NDP on a synthetic dataset consisting of samples from a GP with a Mat\'ern-$\tfrac{3}{2}$ kernel, with a prior on the lengthscale of $\log \c{N}(\log(0.5), \sqrt{0.5})$. In the top row of \cref{fig:kernel-marginalisation} we show samples from this NDP for an increasing number of data points. A sample is coloured according to its most likely lengthscale. As the NDP has no notion of a lengthscale, we infer the most likely lengthscale by retrospectively fitting a set of GPs and matching the lengthscale of the GP corresponding to the highest marginal likelihood to the sample. The bottom row shows the histogram of lengthscales from the produced samples. We observe that the NDP covers the prior at first and then narrows down on the true data-generating lengthscale as more data is observed. Effectively marginalising over the lengthscale posterior.

\subsubsection{Non-Gaussian Posteriors}
\label{sec:experiment:step}

\begin{figure*}[!t]
\centering
\begin{subfigure}{.39\textwidth}
\centering
\includegraphics[width=\textwidth]{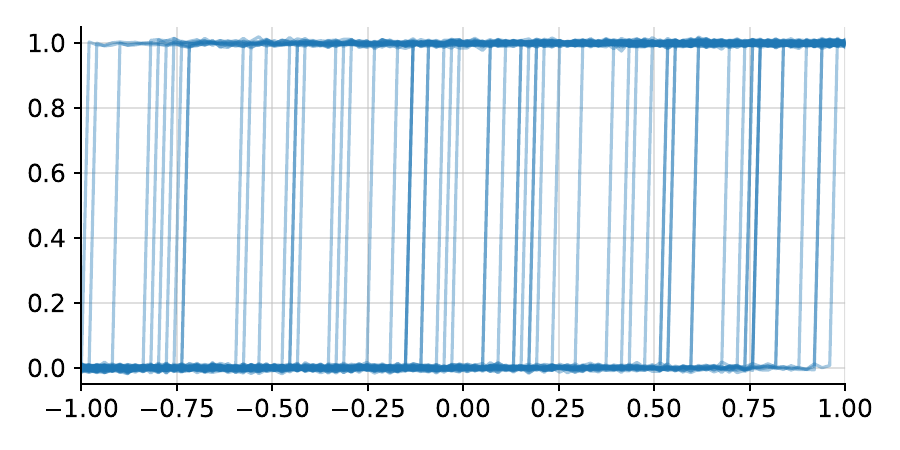}
\caption{Prior\label{fig:step:prior}}
\end{subfigure}
\begin{subfigure}{.39\textwidth}
\centering
\includegraphics[width=\textwidth]{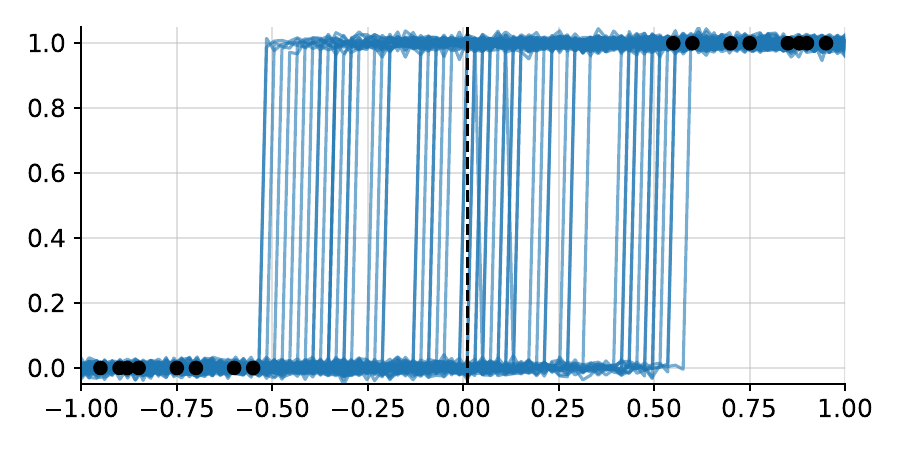}
\caption{Conditional}
\end{subfigure}
\begin{subfigure}{.19\textwidth}
\centering
\includegraphics[width=\textwidth]{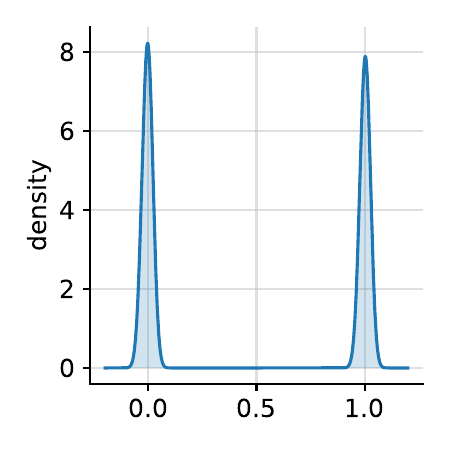}
\caption{$p(y \given x=0.0)$}
\end{subfigure}
\caption{Representing a step function using NDPs. Figure (a) and (b) show samples from the model's prior and conditional distribution, respectively. Figure (c) illustrates the non-Gaussian posterior a NDP can capture, which a GP, by definition, can not do.}
\label{fig:step}
\end{figure*}

The predictions of a GP for a finite set of data points follow a multivariate normal distribution. While this allows for convenient analytic manipulations, it also imposes a restrictive assumption. For example, it is impossible for a GP to represent a 1D step function when the step occurs at a random location within its domain \citep{Neal1998Regression}. In~\cref{fig:step}, we train a NDP on a prior that consists of functions that take a jump at a random location within the interval $[-1, 1]$. Unlike a GP, we observe that the NDP is able to correctly sample from the prior in (a), as well as from the conditional in (b). In (c) we show the a marginal of the NDP's posterior, which correctly captures the bimodal behaviour. This experiment highlights that the NDP can infer a data-driven distribution that need not be Gaussian, which is impossible for GPs.

\begin{figure*}[!bt]
\centering
\begin{subfigure}{.3\textwidth}
\hspace{.1cm}Truth\hspace{.25cm}Context\hspace{.8cm}Samples
\begin{center}
\includegraphics[clip, trim=1cm 0cm 5.5cm 1cm, width=\linewidth]{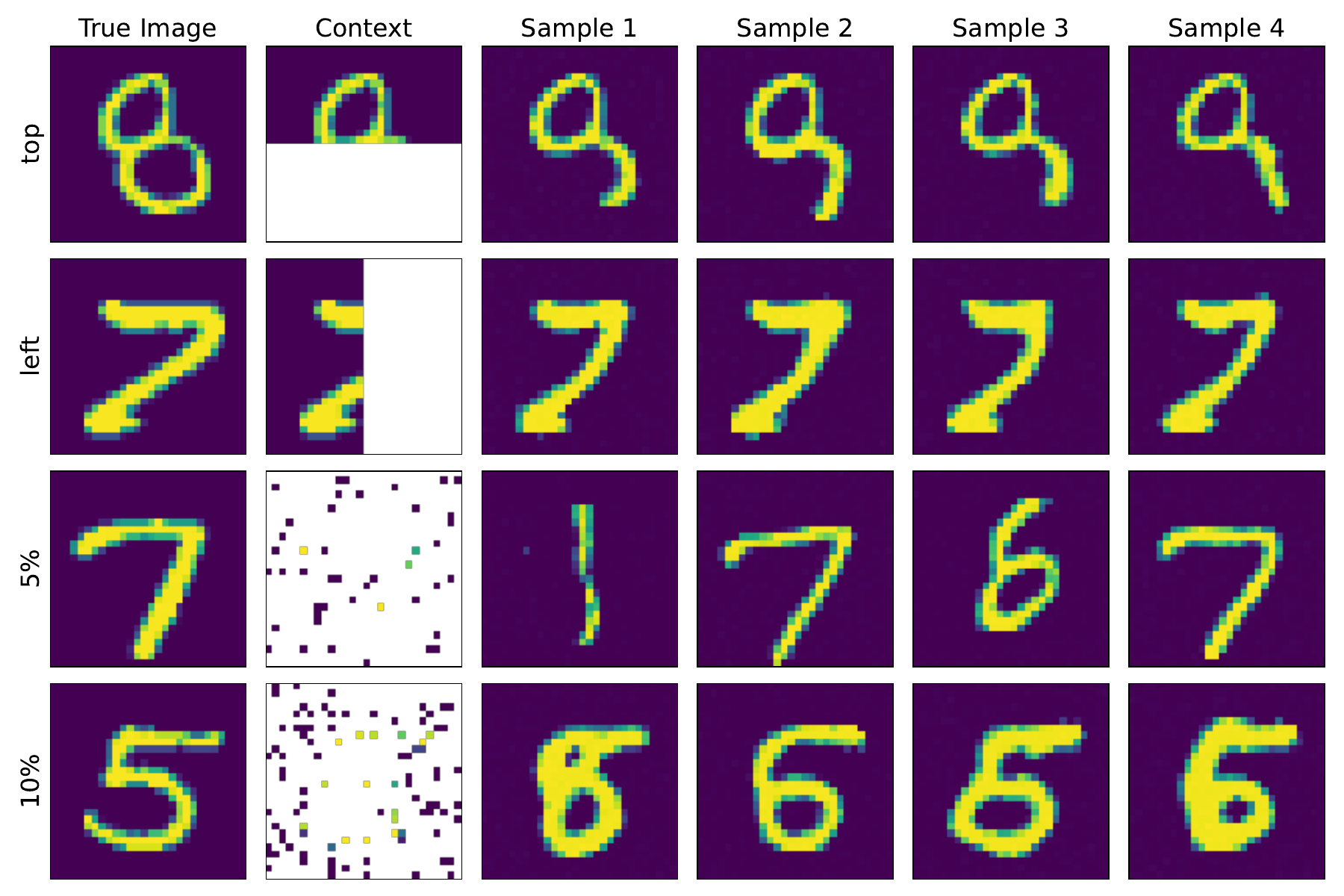}
\label{fig:mnist}
\end{center}
\end{subfigure}
\hspace{5mm}
\begin{subfigure}{.3\textwidth}
\hspace{.1cm}Truth\hspace{.25cm}Context\hspace{.8cm}Samples
\begin{center}
\includegraphics[clip, trim=1cm 0cm 5.5cm 1cm, width=\linewidth]{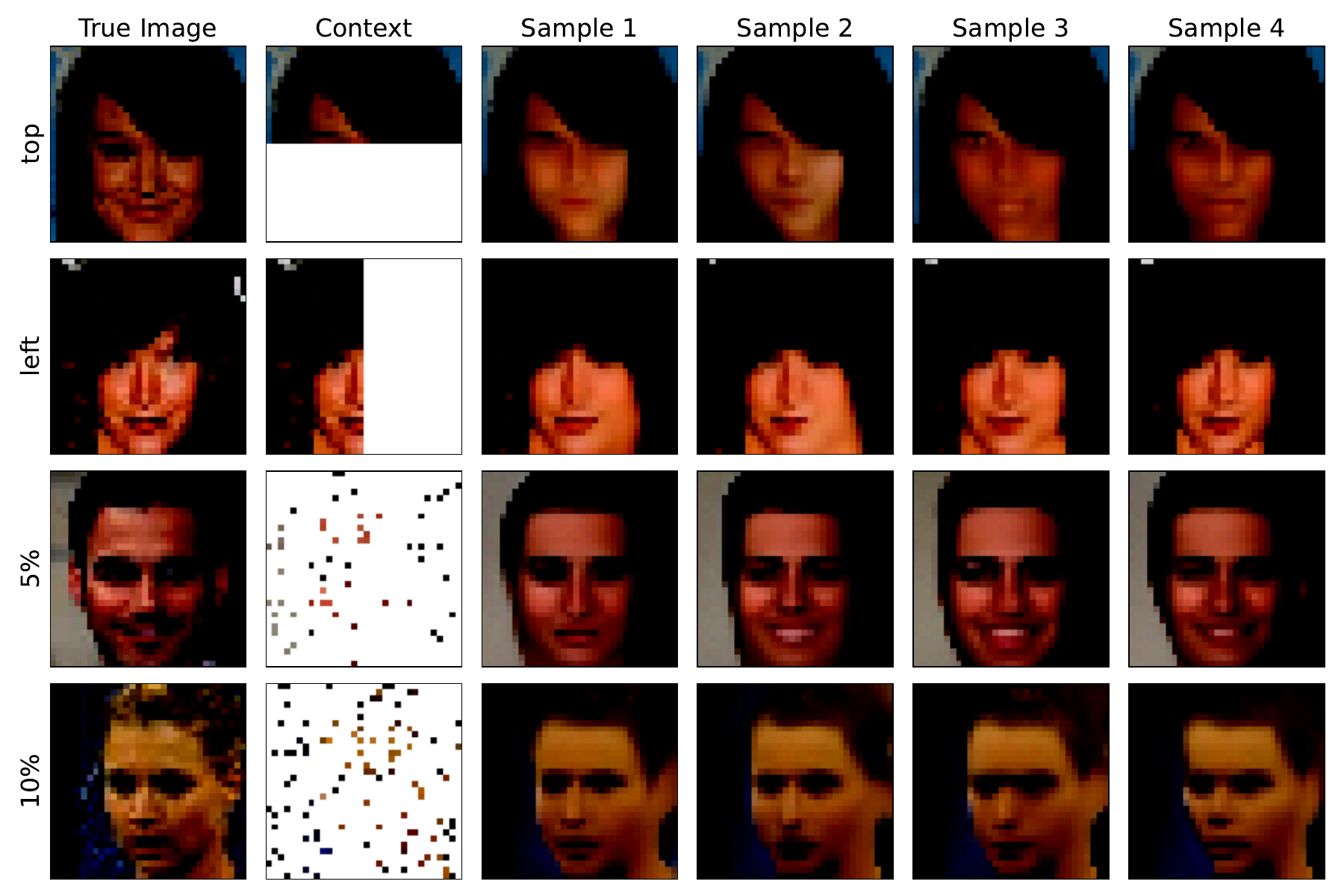}
\label{fig:celeba}
\end{center}
\end{subfigure}
\hspace{3mm}
\begin{subfigure}{.3\textwidth}
\vspace{4mm}
\begin{center}
\includegraphics[width=\linewidth]{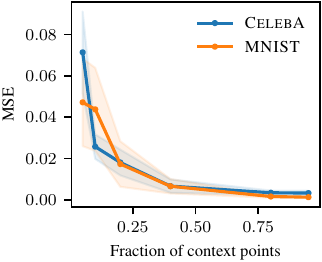}
\label{fig:image-regression-mse}
\end{center}
\end{subfigure}
\caption{NDPs for image regression on {\scshape MNIST} and {\scshape CelebA} ($32\times32$). Figures (a) and (b) show conditional samples where the context datasets are from top to bottom: the upper and left half of the pixels and a random selection of 5\% and 10\% of the pixels. Figure (c) plots the MSE of the NDP's predictions for an increasing number of context points.}
\label{fig:image-regression}
\end{figure*}

\subsubsection{Image regression}
\label{sec:experiment:image-regression}

In the next experiment, we apply NDPs to the task of image regression. The specific goal is to predict pixel values based on their coordinates within the normalized range of $[-2, 2]$. For the {\scshape MNIST} dataset, our task simplifies to predicting a single output value that corresponds to grayscale intensity. However, when tackling the  {\scshape CelebA} $32\times32$ dataset, we deal with the added complexity of predicting three output values for each pixel to represent the RGB colour channels. Crucially, for these image-based tasks, the architecture of our bi-dimensional attention block has been modified to accommodate for the distinct nature of image data in which the order of the input coordinates do matter. This is done by removing the MHSA layer over $D$ in the architecture of \cref{fig:architecture}. Notably, we keep the attention over the input ordering $N$ as this is key when treating images as datasets. 

\Cref{fig:image-regression} shows samples from the NDP conditioned on a variety of contexts (top, left, random 5\% and 10\% of pixels). We observe that the NDP is able to effectively learn the covariance over digits and human faces from the data --- a very challenging task for any non-parametric kernel method. In (c) we measure the MSE of the NDP's predicted output (computed by drawing 5 samples from the conditional and taking the mean) and the target image. We observe an almost perfect match (low MSE) by increasing the context size. The MSE is computed on the normalised pixel values $[0,1]$.

\subsection{Comparison to Neural Processes}
\label{sec:comparison-to-nps}

We now compare NDPs to a range of NP models. 

\begin{table*}[!tb]
\centering
\caption{\label{table:nps}
    Mean test log-likelihood ($\uparrow$) $\pm$ 1 standard error estimated over 128 test samples. Statistically significant best non-GP model is in \textbf{bold}. `--' stands for computationally infeasible models.
}
\begin{tabular}{lcccccc}
\toprule
 & \multicolumn{3}{c}{Squared Exponential} & \multicolumn{3}{c}{Matérn-$\frac52$} \\
 & $D = 1$ & $D = 2$ & $D = 3$ & $D = 1$ & $D = 2$ & $D = 3$ \\
\midrule
{\scshape GP} {(truth)} & $\hphantom{-}0.67 { \pm \scriptstyle 0.03 }$ & $-0.45 { \pm \scriptstyle 0.03 }$ & $-0.94 { \pm \scriptstyle 0.03 }$ & $\hphantom{-}0.19 { \pm \scriptstyle 0.03 }$ & $-0.85 { \pm \scriptstyle 0.03 }$ & $-1.14 { \pm \scriptstyle 0.02 }$ \\
{\scshape NDP} (ours) & $\hphantom{-}\mathbf{0.48} { \pm \scriptstyle 0.04 }$ & $\mathbf{-0.67} { \pm \scriptstyle 0.05 }$ & $\mathbf{-1.16} { \pm \scriptstyle 0.04 }$ & $-0.00 { \pm \scriptstyle 0.04 }$ & $\mathbf{-1.05} { \pm \scriptstyle 0.03 }$ & $\mathbf{-1.33} { \pm \scriptstyle 0.03 }$ \\
\scshape GNP & $\hphantom{-}\mathbf{0.51} { \pm \scriptstyle 0.02 }$ & $-0.98 { \pm \scriptstyle 0.02 }$ & $-1.36 { \pm \scriptstyle 0.02 }$ & $\hphantom{-}\mathbf{0.14} { \pm \scriptstyle 0.02 }$ & $-1.12 { \pm \scriptstyle 0.02 }$ & $-1.37 { \pm \scriptstyle 0.02 }$ \\
\scshape ConvNP & $-0.41 { \pm \scriptstyle 0.06 }$ & $-1.16 { \pm \scriptstyle 0.03 }$ & -- & $-0.63 { \pm \scriptstyle 0.05 }$ & $-1.22 { \pm \scriptstyle 0.02 }$ & -- \\
\scshape ANP & $-0.55 { \pm \scriptstyle 0.05 }$ & $-1.19 { \pm \scriptstyle 0.03 }$ & $-1.36 { \pm \scriptstyle 0.02 }$ & $-0.70 { \pm \scriptstyle 0.04 }$ & $-1.23 { \pm \scriptstyle 0.02 }$ & ${-1.37} { \pm \scriptstyle 0.02 }$ \\
{\scshape GP} (diagonal) & $-0.88 { \pm \scriptstyle 0.07 }$ & $-1.04 { \pm \scriptstyle 0.04 }$ & $-1.22 { \pm \scriptstyle 0.04 }$ & $-0.98 { \pm \scriptstyle 0.06 }$ & $-1.21 { \pm \scriptstyle 0.04 }$ & $-1.31 { \pm \scriptstyle 0.03 }$ \\
\scshape ConvCNP & $-0.74 { \pm \scriptstyle 0.07 }$ & $-1.20 { \pm \scriptstyle 0.03 }$ & $-1.36 { \pm \scriptstyle 0.02 }$ & $-0.87 { \pm \scriptstyle 0.06 }$ & $-1.25 { \pm \scriptstyle 0.03 }$ & $-1.37 { \pm \scriptstyle 0.02 }$ \\
\bottomrule
\end{tabular}
\label{tab:comparison-nps}
\end{table*}

\subsubsection{Regression on Synthetic Data}
\label{sec:experiment:regression}

We evaluate NDPs and NPs on two synthetic datasets, following the experimental setup from \citet{bruinsma2021gaussian} but extending it to multiple input dimensions $D$. We use a Squared Exponential and a Mat\'ern-$\nicefrac{5}{2}$ kernel, where the lengthscale is set to $\ell = \sqrt{D}/4$. We corrupt the samples by white noise $\c{N}(0, 0.05^2)$ and train the models on $2^{14}$ examples. At test time, the context dataset contains between $1$ and $10 \times D$ points, whereas the target set has a fixed size of $50$. We report the test log-likelihood of 128 examples.

For the NDP, the model's likelihood is computed using samples from the conditional distribution of the model. Utilising these samples, the empirical mean and covariance are computed, which are then used to fit a multivariate Gaussian. Given our understanding that the true posterior is Gaussian, this strategy allows for a valid and meaningful comparison with the other methods. Per test example, we used 128 samples to estimate the mean and covariance.

\Cref{tab:comparison-nps} presents the performance of different models: the ground-truth GP, the ground truth GP with diagonal covariance, Gaussian NPs \citep[GNP]{bruinsma2021gaussian}, Convolutional CNPs \citep[ConvCNP]{Gordon2019Convultional}, Convolutional NPs \citep[ConvNP]{foong2020MetaLearning} and Attentive NPs \citep[ANP]{kim2019attentive}. The table demonstrates that for $D=1$, the performance of NDPs is on par with that of GNPs, a model designed specifically to capture Gaussianity. Moreover, both NDPs and GNPs significantly surpass the performance of other methods and nearly reach the level of the ground truth. However, when we scale up to $D=2$ and $3$, NDPs uniquely stand out as the only method that maintains competitiveness while scaling with relative ease.

\subsubsection{Bayesian Optimisation}

\begin{figure*}[!tb]
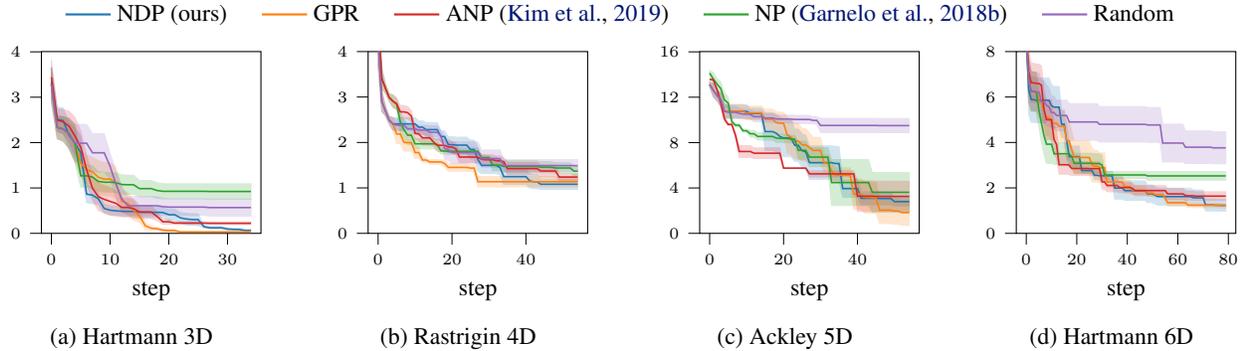

\centering
\begin{subfigure}{\textwidth}
\centering
\begin{tikzpicture}

\definecolor{crimson2143940}{RGB}{214,39,40}
\definecolor{darkgray176}{RGB}{176,176,176}
\definecolor{darkorange25512714}{RGB}{255,127,14}
\definecolor{forestgreen4416044}{RGB}{44,160,44}
\definecolor{steelblue31119180}{RGB}{31,119,180}
\definecolor{mauve}{RGB}{148,103,189}

\begin{axis}[hide axis, xmin={0}, xmax={1}, ymin={0}, ymax={1}, legend columns={-1}, legend style={draw={none}, legend cell align={left}, font={\footnotesize}, /tikz/every even column/.append style={column sep=0.375cm}}]
\addlegendimage{thick,color=steelblue31119180}
\addlegendentry{NDP (ours)}
\addlegendimage{thick,color=darkorange25512714}
\addlegendentry{GPR}
\addlegendimage{thick,color=crimson2143940}
\addlegendentry{ANP \citep{kim2019attentive}}
\addlegendimage{thick,color=forestgreen4416044}
\addlegendentry{NP \citep{garnelo2018neural}}
\addlegendimage{thick,color=mauve}
\addlegendentry{Random}
\end{axis}
\end{tikzpicture}
\end{subfigure}
\begin{subfigure}{.24\textwidth}
\centering
\input{tikz/hartmann3.tex}
\caption{Hartmann 3D}
\label{fig:hartmann3}
\end{subfigure}\hfill
\begin{subfigure}{.24\textwidth}
\centering
\input{tikz/rastrigin4.tex}
\caption{Rastrigin 4D}
\label{fig:rastrigin4}
\end{subfigure}\hfill
\begin{subfigure}{.24\textwidth}
\centering
\input{tikz/ackley5.tex}
\caption{Ackley 5D}
\label{fig:ackley5}
\end{subfigure}\hfill
\begin{subfigure}{.24\textwidth}
\centering
\input{tikz/hartmann6.tex}
\caption{Hartmann 6D}
\label{fig:hartmann6}
\end{subfigure}
\caption{The regret of several probabilistic models used in Thompson sampling-based Bayesian Optimisation.}%
\label{fig:hartmann}
\end{figure*}

We now tackle four black-box optimisation problems in dimensions three to six. The probabilistic models are part of a Bayesian optimisation (BO) loop in which they are used as a surrogate of the expensive black-box objectives. At each iteration, the surrogate is evaluated at 128 random locations in the input domain, and the input corresponding to the minimum value is selected as the next query point. The objective is evaluated at this location and added to the context dataset (akin to Thompson sampling BO \citep{shahriari2015taking}). \Cref{fig:hartmann} shows the regret (distance from the true minimum) for the different models. We observe that the NDP almost matches the performance of GPR, which is the gold standard model for this type of task. NDPs also outperform the NPs and random search strategies. The important difference between GPR and the NDP is that the NDP requires \emph{no training} during the BO loop, whereas the GPR is retrained at every step.

\subsection{Modelling Function Inputs and Outputs Jointly}
\label{sec:experiment:joint}

\begin{figure*}[!tb]
\centering
\begin{subfigure}{\textwidth}
\centering
\includegraphics{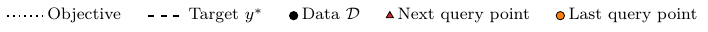}
\end{subfigure}
\begin{subfigure}{\textwidth}
\centering
\includegraphics[width=.88\textwidth]{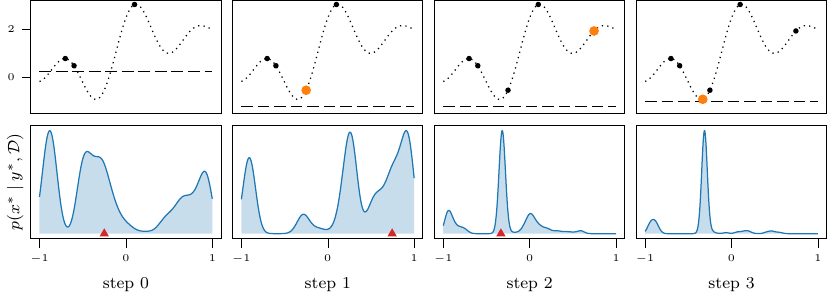}
\end{subfigure}
\caption{Global optimisation by diffusing the input locations. We condition the model at each step on a target $y^*$ value (dashed line), and use the NDP to sample from $p(x^* \given y^*, \c{D})$. The panels in the upper row illustrate the progression of the optimisation. Bottom row shows the distribution $p(x^* \given y^*, \c{D})$ and the red triangles mark the next selected query point.}
\label{fig:bo1d}
\end{figure*}
So far, we have used NDPs to model $p(\v{y} \given \v{x}, \c{D})$. This is a natural choice as in regression it is typically the only quantity of interest to make predictions. However, we can extend NDPs to model the joint $p(\v{x}, \v{y} \given \c{D})$. For this, during the forward process we corrupt both the function inputs and outputs with additive Gaussian noise. The task of the reverse process now consists of denoising both the corrupted inputs $\v{x}_t$ and outputs $\v{y}_t$, which leads to the objective
\begin{multline}
\label{eq:objective-full}
\c{L}_\theta = \E_{t, \v{x}_0, \v{y}_0, {\color{darkorange25512714}\v{\varepsilon}_x}, \v{\varepsilon}_y} \Big[ {\color{darkorange25512714}\|\v{\varepsilon}_x - \v{\epsilon}^x_\theta(\v{x}_t, \v{y}_t, t)\|^2}
+ \\
\|\v{\varepsilon}_y - \v{\epsilon}^y_\theta(\v{x}_{\color{darkorange25512714}t},  \v{y}_t, t)\|^2
\Big],
\end{multline}
where we highlighted the differences with \cref{eq:objective} in {\color{darkorange25512714}orange}. Importantly, in this case we require a noise model for both the inputs and outputs: $\v{\epsilon}_\theta^x$ and $\v{\epsilon}_\theta^y$, resp. We detail the architecture of this NN in the supplementary (\cref{fig:architecture-full}). We design it such that $\v{\epsilon}_\theta^x$ and $\v{\epsilon}_\theta^y$ share most of the weights apart from the final invariance layers as explained in \cref{sec:architecture}. The inputs to the NN are now the corrupted inputs $\v{x}_t = \sqrt{\bar{\alpha}_t} \v{x}_0 + \sqrt{1-\bar{\alpha}_t} \v{\varepsilon}_x$ and outputs $\v{y}_t = \sqrt{\bar{\alpha}_t} \v{y}_0 + \sqrt{1-\bar{\alpha}_t} \v{\varepsilon}_y$. This is in contrast to the previous NDP in which $\v{x}_t = \v{x}_0$ for all $t$.

\subsubsection{Global optimisation using NDPs}
\label{sec:global-optimisation}

By taking advantage of the NDP's ability to model the full joint distribution $p(\v{x}, \v{y} \given \c{D})$ we conceive of a new global optimisation strategy. Consider conditioning the NDP on the current belief about the minima $y^*$. A NDP allows us to obtain samples from $p({x}^* \given {y}^*)$ which provides information about where the minima lie in the input domain. In \cref{fig:bo1d} we illustrate a global optimisation routine using this idea. In each step, we sample a target from $p(y^*)$, which describes our belief about the minima. For the experiment, we sample $y^*$ from a truncated-normal, where the mean corresponds to the minimum of the observed function values and the variance corresponds to the variance of the observations. The next query point is selected by sampling $p({x}^* \given {y}^*, \c{D})$, thereby systematically seeking out the global minimum. This experiment showcases NDP's ability to model the complex interaction between inputs and outputs of a function ---a task on which it was not trained.

\section{Conclusion}

\label{sec:conclusions}
We proposed Neural Diffusion Processes (NDPs), a denoising diffusion generative model for learning distribution over functions, and generating prior and conditional samples. NDPs generalise diffusion models to infinite-dimensional functions by allowing the indexing of the function's marginals. We introduced the bi-dimensional attention block, which wires dimension and sequence equivariance into the architecture such that it satisfies the basic properties of a stochastic process. We empirically show that NDPs are able to capture functional distributions that are richer than Gaussian processes, and more accurate than neural processes. We concluded the paper by proposing a novel optimisation strategy based on diffusing the inputs. The code is available at \url{https://github.com/vdutor/neural-diffusion-processes}.

\paragraph{Limitations}
As with other diffusion models, we found that the sample quality improves with the number of diffusion steps. This does however lead to slower inference times. Techniques for accelerating the inference process could be incorporated to ameliorate this issue. Secondly, as is common with NNs we found that it is important for test input points to lie within the training range, as going beyond leads to poor performance.

\section*{Acknowledgements}
We would like to recognise the contributions of Carl Henrik Ek, Erik Bodin, Sebastian Ober, and anonymous reviewers for their insightful feedback and constructive suggestions. We're also thankful to Yeh Whye Teh, Emile Mathieu, and Michael Hutchinson for their valuable insights, particularly concerning the topic of consistency. Their inputs have substantially shaped the presentation of this work.

\bibliography{citations}
\bibliographystyle{icml2023}

\newpage
\appendix
\onecolumn

\section{Derivation of the Loss}
\label{sec:app:derivation-lower-bound}

In this section we derive the objectives given in \cref{eq:objective-full,eq:objective}, following a path similar to that presented in \citet{ho2020denoising}. Let $\v{s}_t = (\v{x}_t, \v{y}_t)$ be the state that combines both the inputs and observations. We wish to find the set of parameters $\theta$ which maximise the likelihood given the initial state, $p_\theta(\v{s}_0)$. While a direct evaluation of the likelihood appears intractable,
\begin{equation*}
     p_\theta(\v{s}_0) = \int p_\theta(\v{s}_{0 \ldots T}) d \v{s}_{1 \ldots T} \, ,
\end{equation*}
this may be recast in a form which allows for a comparison to be drawn between the forward and reverse trajectories \citep{sohl2015deep}
\begin{equation*}
     p_\theta(\v{s}_0) = \int  p_\theta(\v{s}_T) q(\v{s}_{1 \ldots T}|\v{s}_0)  \prod_{t=1}^{T} \frac{p_\theta(\v{s}_{t-1}|t)}{q(\v{s}_{t} | \v{s}_{t-1})} d \v{s}_{1 \ldots T} \, .
\end{equation*}

The appeal of introducing the reverse process is that it is tractable when conditioned on the first state $\v{s}_0$, taking a Gaussian form
\begin{equation*}
    q(\v{s}_{t-1} \given \v{s}_{t}, \v{s}_{0}) = \c{N}(\v{s}_{t-1} \given \tilde{\v{\mu}}(\v{s}_t, \v{s}_0), \tilde{\beta}_t \m{I}).
\end{equation*}

Our loss function reflects a lower bound on the negative log likelihood 
\begin{equation*}
    \E[-\log p_\theta(\v{s}_0)]   \leq \E_{q(\v{s}_{0:T})}\left[ \log \frac{q(\v{s}_{1:T} \given \v{s}_0)}{p_\theta(\v{s}_{0:T})} \right] := \c{L}_\theta   \, ,
\end{equation*}
which may be decomposed into two edge terms and a sum over the intermediate steps, as follows 

\begin{equation*}
    \c{L}_{\theta} = \E_q \left[ L_T + \sum_{t=2}^T L_{t-1}  + L_0  \right] \, ,
\end{equation*}
where
\begin{equation}
\begin{split}
L_0 &= - \log p_\theta(\v{s}_0 \given \v{s}_1) \, , \\
L_{t-1} &= \textrm{KL}(q(\v{s}_{t-1} \given \v{s}_t, \v{s}_0) \| p_\theta(\v{s}_{t-1} \given \v{s}_{t})) \, , \\
L_T &= \textrm{KL}(q(\v{s}_T \given \v{s}_0) \| p_\theta(\v{s}_T)) \, .
\end{split}
\end{equation}

We can write 

\begin{equation} \label{eq:lt}
L_{t-1} = \E_{\v{s}_0, \v{\varepsilon}} \left[ \frac{1}{2 \sigma^2} \| \tilde{\v{\mu}}(\v{s}_t, \v{s}_0) - \v{\mu}_\theta(\v{s}_t, t) \|^2 \right] \, ,
\end{equation}
where the mean is a function of the previous and first state
\begin{equation*}
    \tilde{\v{\mu}}(\v{s}_t, \v{s}_0) = \frac{\sqrt{\alpha_t} (1 - \bar{\alpha}_{t-1})}{1 - \bar{\alpha}_{t}} \v{s}_{t} + \frac{\sqrt{\bar{\alpha}_{t-1}} \beta_t}{1-\bar{\alpha}_t} \v{s}_0
\end{equation*}
and the variance
\begin{equation*}
  \tilde{\beta}_t =  \frac{1 - \bar{\alpha}_{t-1}}{1-\bar{\alpha}_t} \beta_t \, .
\end{equation*}

It is helpful to consider the relationship between the initial and final states

\begin{equation*}
    \v{s}_0 = \frac{1}{\sqrt{\bar{\alpha}_t}}(\v{s}_t - \sqrt{1 - \bar{\alpha}_t} \v{\varepsilon}_s),
\end{equation*}
where $\v{\varepsilon}_s = (\v{\varepsilon}_x,\v{\varepsilon}_y)$, so that we can rewrite the mean as
\begin{equation}
\label{eq:mu_tilde}
    \tilde{\v{\mu}}(\v{s}_t, \v{\varepsilon}_s) = \frac{1}{\sqrt{\alpha_t}}(\v{s}_t - \frac{\beta_t}{\sqrt{1-\bar{\alpha}_t}}\v{\varepsilon}_s) \, .
\end{equation}

Equation \ref{eq:lt} can now be expressed as

\begin{equation}
L_{t-1} = \E_{\v{s}_0, \v{\varepsilon}_s} \left[ \frac{\beta_t^2}{2 \sigma^2 \alpha_t (1 - \bar{\alpha}_t)} \| \v{\varepsilon}_s - \v{\epsilon}^s_{\theta} \|^2 \right] \, .
\end{equation}

\begin{figure}[t]
    \centering
    \includegraphics[width=\textwidth]{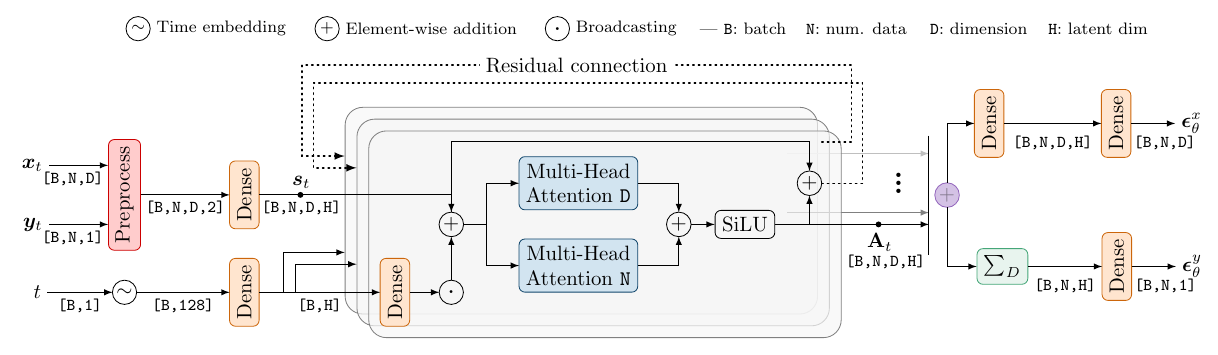}
    \caption{Architecture of the NDP's NN noise models $\v{\epsilon}_\theta^x$ and $\v{\epsilon}_\theta^y$. Compared to \cref{fig:architecture} this architecture has two outputs, one to predict the corruption on the inputs $\v{x}_t$ and one to predict the corruption on the function outputs $\v{y}_t$. Both output share a lot of weights in the network and only bifurcate in the last few layers.}
    \label{fig:architecture-full}
\end{figure}

Finally, since the variance schedule is fixed, and the edge terms are not found to improve empirical performance, our simplified training objective is given by 

\begin{equation}
\label{eq:objective_appendix}
    \c{L}_\theta = \E_{\v{x}_0, \v{y}_0, \v{\varepsilon}_x, \v{\varepsilon}_y, t}\Big[ \|\v{\varepsilon}_x - \v{\epsilon}^x_{\theta}(\v{x}_t, \v{y}_t, t)\|^2 + \|\v{\varepsilon}_y - \v{\epsilon}^y_{\theta}(\v{x}_t, \v{y}_t, t)\|^2 \Big].
\end{equation}
So far we have derived the objective of the `full' NDP model (i.e., the model which diffuses both $\v{x}$ and $\v{y}$) given in \cref{eq:objective-full} of the main paper. In \cref{fig:architecture-full} we detail the architecture for the noise models $\v{\epsilon}_\theta^x(\cdot)$ and $\v{\epsilon}_\theta^y(\cdot)$.

\section{Algorithms}

In this section we list pseudo-code for training and sampling NDPs.

\label{sec:app:pseudo-code}
\subsection{Training}

\begin{algorithm}[H]
\centering
\caption{Training}\label{alg:training}
\begin{algorithmic}
\INPUT Input distribution $q(\v{x}_0)$ and covariance function $k_{\psi}$, with prior over hyperparameters $p_\psi$. Noise schedule $\beta_t$ for $t \in \{1, 2, \ldots, T\}$. A loss function $L$ (e.g., MSE or MAE).

\vspace{2mm}
\hspace{-3.5mm}\textbf{begin}
\STATE Precompute $\gamma_t = \sqrt{1-\bar{\alpha}_t}$ and $\bar{\alpha}_t = \prod_{j=1}^t (1 - \beta_j)$.
\FOR{$i=1,2,\cdots,N_{\mathrm{iter}}$}
\STATE Sample $\v{x}_0\sim q(\v{x}_0)$, $\psi \sim p_{\psi}$, $\v{y}_0 \sim \c{N}(\v{0}, k_{\psi}(\v{x}_0, \v{x}_0) + \sigma^2 \m{I})$.
\STATE Sample $\v{\varepsilon} \sim\c{N}(\v{0},\m{I})$, and $t\sim\c{U}(\{1,\ldots,T\})$.
\STATE Compute $\v{y}_t = \sqrt{\bar{\alpha}_t} \v{y}_0 + \gamma_t \v{\varepsilon}$.\hfill 
\STATE Update $\theta$ using gradient $\nabla_{\theta} L(\v{\varepsilon}, \v{\epsilon}_{\theta}({\v{x}_0}, \v{y}_t, t)) $. 
\ENDFOR
\end{algorithmic}
\end{algorithm}

\subsection{Prior and Conditional Sampling}
\label{sec:app:code-conditional-sampling}

In practice, the code to sample the prior is a special case of the conditional code with an empty context dataset and $U=1$. However, here we list them both for the clarity of the exposition. %

\begin{algorithm}[H]
\centering
\caption{Prior Sampling}\label{alg:prior-sampling}
\begin{algorithmic}
\INPUT  A set of input locations $\v{x}_0$, a NDP noise model $\v{\epsilon}_{\theta}(\cdot)$.  $\gamma_t$ and $\bar{\alpha}_t$ for a given noise schedule $\beta_t$.

\vspace{2mm}

\hspace{-3.5mm}\textbf{begin}
\STATE Sample a random initial state and $\v{y}_T$ from $\c{N}(\v{0},\m{I})$.
\FOR{$t=T,T-1,\ldots,1$}
\STATE Sample using backward kernel:
    $$\v{y}_{t-1} \sim
         \c{N}\big(\frac{1}{\sqrt{1 - \beta_t}} ({\v{y}_t} - \frac{\beta_t}{\gamma_t} {\v{\epsilon}_{\theta}(\v{x}_0, \v{y}_t, t)}), \frac{\gamma^2_t}{\gamma^2_{t-1}} \beta_t\m{I}\big).$$
\ENDFOR
\OUTPUT $(\v{x}_0, \v{y}_0)$
\end{algorithmic}
\end{algorithm}

\begin{algorithm}[H]
\centering
\caption{Conditional Sampling}\label{alg:conditional-sampling}
\begin{algorithmic}
\INPUT A context dataset $\c{D} = \{([\v{x}_0^c]_i \in \c{X}, [\v{y}_0^c]_i \in \R)\}_{i=1}^N$. 
A NDP noise model $\v{\epsilon}_{\theta}(\cdot)$. $\gamma_t$ and $\bar{\alpha}_t$ for a given noise schedule $\beta_t$. 

\vspace{2mm}
\hspace{-3.5mm}\textbf{begin}

\STATE Sample ${\v{y}}^*_T \sim \c{N}(\v{0},\m{I})$.
\STATE Let ${\v{x}}_0 = [\v{x}_0^c, \v{x}_0^*]$

\FOR{$t=T,T-1,\ldots,1$}
\FOR{$u=1,\ldots,U$}
\STATE Sample context points t steps forward
\begin{equation}
\v{y}_t^c \~ \c{N}\Big(
\sqrt{\bar{\alpha}_t} \v{y}_0^c, (1-\bar{\alpha}_t)\,\m{I}
\Big).
\end{equation}
\STATE Let $\v{y}_t =[\v{y}_t^c, \v{y}^*_t]$.
\STATE Sample backward
        \hfill\cref{eq:backward}
    $$\v{y}_{t-1} \sim
         \c{N}\big(\frac{1}{\sqrt{1 - \beta_t}} ({\v{y}_t} - \frac{\beta_t}{\gamma_t} {\v{\epsilon}_{\theta}(\v{x}_0, \v{y}_t, t)}), \frac{\gamma^2_t}{\gamma^2_{t-1}} \beta_t\m{I}\big).$$
\STATE Diffuse forward by one step
$$
\v{y}_t \sim \c{N}\big(\sqrt{1-\beta_{t}} \v{y}_{t-1}, \beta_{t}\m{I} \big)
$$
\ENDFOR
\ENDFOR
\OUTPUT{$(\v{x}_0, \v{y}_0)$}
\end{algorithmic}
\end{algorithm}

\section{Proofs}
\label{sec:app:neural-network-properties}

In this section, we shall formally demonstrate that NDP's noise model adheres to the symmetries associated with permutations of the datapoint orderings and the permutations of the input dimensions. We focus our attention to the full noise model of \cref{fig:architecture-full} because \cref{fig:architecture} is a simplification of it in which we only keep a single output. We start by proving properties of its main building block: the bi-dimensional attention block. We can then straightforwardly prove the equivariance and invariance properties that hold for the NDP's noise model. Before that, we start by setting the notation.

\subsection{Notation, Definitions and Preliminary lemmas}

\renewcommand{\eqnannotationtext}[1]{\footnotesize#1\strut}
\newcommand{\attnN}{\ensuremath{\textrm{Attn}_N}\xspace}
\newcommand{\attnD}{\ensuremath{\textrm{Attn}_D}\xspace}

\paragraph{Notation.}
Let $\v{s} \in \R^{N \times D \times H}$ be a tensor of rank (or dimension) three, where we refer to each dimension according to the following convention:
\begin{equation*}
    \textrm{shape}(\v{s}) = [N, D, H],
\end{equation*}
where $N$ stands for the sequence length, $D$ the input dimensionality and $H$ the embedding.

In the next definitions we will use NumPy-based indexing and slicing notation. We assume the reader is familiar with this convention. Most notably, we use a colon ($:$) to reference every element in a dimension.

\begin{definition}
    Let ${\Pi}_{N}$ be the set of all permutations of indices $\{1, \ldots, N\}$. Let $\pi_n \in \Pi_N$ and $\v{s} \in \R^{N \times D \times H}$. Then $(\pi_n \circ \v{s}) \in \R^{N \times D \times H}$ denotes a tensor where the ordering of indices in the first dimension are reshuffled (i.e.,~permuted) according to $\pi_n$. We write
    \begin{equation*}
        \pi_n \circ \v{s} = \v{s}_{\pi_n(1), \pi_n(2),\ldots,\pi_n(N);\,:\,;\,:}
    \end{equation*}
\end{definition}

\begin{definition}
    Let ${\Pi}_{D}$ be the set of all permutations of indices $\{1, \ldots, D\}$. Let $\pi_d \in \Pi_D$ and $\v{s} \in \R^{N \times D \times H}$. Then $(\pi_d \circ \v{s}) \in \R^{N \times D \times H}$ denotes a tensor where the ordering of indices in the second dimension are reshuffled (i.e.,~permuted) according to $\pi_d$. We write
    \begin{equation*}
        \pi_d \circ \v{s} = \v{s}_{:\,;\,\pi_d(1), \pi_d(2),\ldots,\pi_d(D);\,:}.
    \end{equation*}
\end{definition}

\begin{definition}
    A function $f:\R^{N \times D \times H} \rightarrow \R^{N \times D \times H}$ is \textit{equivariant to $\Pi$} if for any permutation $\pi \in \Pi$, we have
    \begin{equation*}
        f(\pi \circ \v{s}) = \pi \circ f(\v{s}).
    \end{equation*}
\end{definition}

\begin{definition}
    A function $f:\R^{N \times D \times H} \rightarrow \R^{N \times D \times H}$ is \textit{invariant to $\Pi$} if for any permutation $\pi \in \Pi$, we have
    \begin{equation*}
        f(\pi \circ \v{s}) = f(\v{s}).
    \end{equation*}
\end{definition}
Invariance in layman's terms means that the output is not affected by a permutation of the inputs.

\begin{lemma}
    \label{lemma:composition}
    The composition of equivariant functions is equivariant.
    \end{lemma}
    \begin{proof}
    Let $f$ and $g$ be equivariant to $\Pi$, then for all $\pi \in \Pi$:
    \begin{equation*}
        (f \circ g)(\pi \circ \v{s}) = f ( g(\pi \circ \v{s})) = f ( \pi \circ g(\v{s})) = \pi \circ f(g(\v{s})),
    \end{equation*}
    which shows that the composition, $f \circ g$, is equivariant as well.
    \end{proof}

\begin{lemma}
    \label{lemma:elementwise}
    An element-wise operation between equivariant functions remains equivariant.
\end{lemma}
\begin{proof}
    Let $f$ and $g$ be equivariant to $\Pi$, then for all $\pi \in \Pi$ and an element-wise operation $\oplus$ (e.g.,~addition), we have
    \begin{equation*}
    (f \oplus g)(\pi \circ \v{s}) = f(\pi \circ \v{s}) \oplus g(\pi \circ \v{s}) =
    (\pi \circ f(\v{s})) \oplus (\pi \circ g(\v{s})) = \pi \circ (f \oplus g).
    \end{equation*}
    which shows that the composition, $f \oplus g$, is equivariant as well.
\end{proof}

\begin{lemma}
\label{lemma:row-equivariance}
A function $f:\R^{N \times D \times H} \rightarrow \R^{N \times D \times H}$ which applies the same function $g$ to all its rows, i.e. $f: \v{s} \mapsto [g(\v{s}_{1;:;:}), g(\v{s}_{2;:;:}),\ldots, g(\v{s}_{N;:;:})]$ with $g:\R^{D \times H} \rightarrow \R^{D \times H}$ is equivariant in the first dimension.
\end{lemma}
\begin{proof}
    Follows immediately from the structure of $f$.
\end{proof}

\subsection{Bi-Dimensional Attention Block}
\label{sec:app:proof-bi-dimensional-attention-block}

With the notation, definitions and lemmas in place, we now prove that the bi-dimensional attention block is equivariant in its first and second dimension, respectively to permutations in the set $\Pi_N$ and $\Pi_D$. We start this section by formally defining the bi-dimensional attention block and its main components attention components \attnN and \attnD. Finally, in \cref{sec:app:neural-network-properties-final} we prove the properties of the noise models $\v{\epsilon}_{\theta}^x$ and $\v{\epsilon}_{\theta}^y$ making use of the results in this section.

\begin{definition}
Ignoring the batch dimension $B$, let $\m{A}_t: \R^{N \times D \times H} \rightarrow \R^{N \times D \times H}; \v{s} \mapsto \m{A}_t(\v{s})$ be the bi-dimensional attention block. As illustrated in \cref{fig:architecture-full}, it operates on three dimensional tensors in $\R^{N\times D \times H}$ and applies attention across the first and second dimension using \attnN and \attnD, respectively. The final output $\m{A}_t(\v{s})$ is obtained by summing the two attention outputs, before applying an element-wise non-linearity.
\end{definition}

We now proceed by defining the component \attnD and its properties. Subsequently, in \cref{def:attnN} we define \attnN and its properties. Finally, we combine both to prove \cref{prop:one} in the main paper about the bi-dimensional attention block.

\begin{definition}
    \label{def:attnD}
    Let $\attnD: \R^{N \times D \times H} \rightarrow \R^{N \times D \times H}$ be a self-attention block \citep{vaswani2017attention} acting across $D$. 
    Let $\sigma$ be a softmax activation function operating on the last dimension of a tensor. Then, \attnD is defined as
    \begin{equation*}
        \attnD(\v{s})[n, d, h] = 
        \sum_{d'=1}^D \m{\Sigma}_{n,d,d'}(\v{s}) \v{s}^v_{n, d', h},
    \quad\text{where}\quad
        \m{\Sigma}_{n,d,d'}(\v{s}) = \sigma\big(\frac{1}{\sqrt{H}} \sum_l \v{s}^k_{n,d,\ell} \v{s}^q_{n,d',\ell}\big) 
    \end{equation*}
    given a linear projection of the inputs $\v{s}$ which maps them into keys ($k$), queries ($q$) and values ($v$)
    \begin{equation*}
        \v{s}^k_{n,d,\ell} = \sum_j \v{s}_{n,d,j} \m{W}^k_{j,\ell},\quad
        \v{s}^q_{n,d',\ell} = \sum_j \v{s}_{n,d,j} \m{W}^q_{j,\ell},\quad
        \v{s}^v_{n, d', h} = \sum_j \v{s}_{n,d,j} \m{W}^v_{j,\ell}.
    \end{equation*}
\end{definition}

\begin{proposition}
    \label{prop:attnD}
    \attnD is equivariant to $\Pi_N$ and $\Pi_D$ (i.e.,~across sequence length and input dimensionality).
\end{proposition}

\begin{proof} We prove the equivariance to $\Pi_N$ and $\Pi_D$ separately.
First, from the definition we can see that \attnD is a function that acts on each element row of $\v{s}$ separately. Thus, by \cref{lemma:row-equivariance}, \attnD is equivariant to $\Pi_N$.
    
Next, we want to prove equivariance to $\Pi_D$. We want to show that for all $\pi_d \in \Pi_D$:
\begin{equation*}
    \attnD(\pi_d \circ \v{s}) =  \pi_d \circ \attnD(\v{s}).
\end{equation*}
The self-attention mechanism consists of a matrix multiplication of the attention matrix $\Sigma$ and the projected inputs. We start by showing that the attention matrix is equivariant to permutations in $\Pi_D$
\begin{align*}
    \m{\Sigma}_{n,d,d'}(\pi_d \circ \v{s})  
    &=  \sigma\Big( \frac{1}{\sqrt{H}} \sum_\ell \big(\pi_d \circ \v{s}\big)^k_{n,d,\ell}\big(\pi_d \circ \v{s}\big)^q_{n,d,\ell} \Big) \\
    &=\sigma\Big(
        \frac{1}{\sqrt{H}} \sum_{\ell, j} \big( \v{s}_{n,\pi_d(d),j}\m{W}^k_{j,\ell}\big)\big(\v{s}_{n,\pi_d(d'),j}\m{W}^q_{j,\ell}
        \big) \Big) \\
    &= \m{\Sigma}_{n,\pi_d(d),\pi_d(d')}(\v{s})
\end{align*}
It remains to show that the final matrix multiplication step restores the row-equivariance
\begin{align*}
    \attnD(\pi_d \circ \v{s}) 
    &= \sum_{d'}  \m{\Sigma}_{n,d,d'}(\pi_d \circ \v{s}) \big( \pi_d \circ \v{s}\big)^v_{n,d',h}\\
    &= \sum_{d'}  \m{\Sigma}_{n,\pi_d(d),\pi_d(d')}(\v{s}) \v{s}^v_{n,\pi_d(d'),h} \\
    &= \sum_{d'}  \m{\Sigma}_{n,\pi_d(d),d'}(\v{s}) \v{s}^v_{n,d',h}
    = \pi_d \circ \attnD(\v{s}).
\end{align*}
This concludes the proof.
\end{proof}

We continue by defining and proving the properties of the second main component of the bi-dimensional attention block: \attnN.

\begin{definition}
    \label{def:attnN}
    Let $\attnN: \R^{N \times D \times H} \rightarrow \R^{N \times D \times H}$ be a self-attention block \citep{vaswani2017attention} acting across $N$ (i.e. the sequence length). 
    Let $\sigma$ be a softmax activation function operating on the last dimension of a tensor. Then \attnN is defined as
    \begin{equation*}
        \attnN(\v{s})[n, d, h] = 
        \sum_{n'=1}^N \m{\Sigma}_{n,d,n'}(\v{s}) \v{s}^v_{n', d, h},
    \text{ where}\quad
        \m{\Sigma}_{n,d,n'}(\v{s}) = \sigma\big(\frac{1}{\sqrt{H}} \sum_l \v{s}^k_{n,d,\ell} \v{s}^q_{n',d,\ell}\big) 
    \end{equation*}
    given a linear projection of the inputs $\v{s}$ which maps them into keys ($k$), queries ($q$) and values ($v$)
    \begin{equation*}
        \v{s}^k_{n,d,\ell} = \sum_j \v{s}_{n,d,j} \m{W}^k_{j,\ell},\quad
        \v{s}^q_{n',d,\ell} = \sum_j \v{s}_{n',d,j} \m{W}^q_{j,\ell},\quad
        \v{s}^v_{n', d, h} = \sum_j \v{s}_{n',d,j} \m{W}^v_{j,\ell}.
    \end{equation*}
\end{definition}

\begin{proposition}
    \attnN is equivariant to $\Pi_N$ and $\Pi_D$ (i.e.,~across sequence length and input dimensionality).
\end{proposition}

\begin{proof} 
    Follows directly from \cref{prop:attnD} after transposing the first and second dimension of the input.
\end{proof}

Finally, we have the necessary ingredients to prove \cref{prop:one} from the main paper.

\begin{proposition}
\label{prop:attention-block}
    The bi-dimensional attention block $\m{A}_t$ is equivariant to $\Pi_D$ and $\Pi_N$.
\end{proposition}

\begin{proof}
    The bi-dimensional attention block simply adds the output of $\attnD$ and $\attnN$, followed by an element-wise non-linearity. Therefore, as a direct application of \cref{lemma:elementwise} the complete bi-dimensional attention block remains equivariant to $\Pi_N$ and $\Pi_D$.
\end{proof}

\subsection{NDP Noise Model}
\label{sec:app:neural-network-properties-final}

By building on the equivariant properties of the bi-dimensional attention block, we prove the equivariance and invariance of the NDP's noise models, denoted by $\v{\epsilon}^x_{\theta}$ and $\v{\epsilon}^y_{\theta}$, respectively. We refer to \cref{fig:architecture-full} for their definition. In short, $\v{\epsilon}^x_{\theta}$  consists of adding the output of several bi-dimensional attention blocks followed by dense layers operating on the last dimension. Similarly, $\v{\epsilon}^y_{\theta}$ is constructed by summing the bi-dimensional attention blocks, but is followed by a summation over $D$ before applying a final dense layer.

The following propositions hold:

\begin{proposition}
    The function $\v{\epsilon}^x_{\theta}$ is equivariant to $\Pi_D$ and $\Pi_N$.
\end{proposition}

\begin{proof}
    The output $\v{\epsilon}^x_{\theta}$ is formed by element-wise summing the output of bi-dimensional attention layers. Directly applying \cref{lemma:composition} and \cref{prop:attention-block} completes the proof.
\end{proof}

\begin{proposition}
    The function $\v{\epsilon}^y_{\theta}$ is equivariant to $\Pi_N$.
\end{proposition}

\begin{proof}
    The summation over $D$ does not affect the equivariance over $\Pi_N$ from the bi-dimensional attention blocks as it can be cast as a row-wise operation.
\end{proof}

\begin{proposition}
    The function $\v{\epsilon}^y_{\theta}$ is invariant to $\Pi_D$.
\end{proposition}

\begin{proof}
    Follows from the equivariance of the bi-dimensional blocks and \citep[Thm. 7]{zaheer2017deep}.
\end{proof}

\section{A Primer on Neural Processes}
\label{sec:app:nps}

While Neural Diffusion Processes (NDPs) and Neural Processes (NPs) share a similar goal, they differ significantly in their approach. 

NPs learn a function that maps inputs $x \in \R^d$ to outputs $y \in \R$.  NPs specifically define a family of conditional distributions that allows one to model an arbitrary number of targets $\v{x}^*$ by conditioning on an arbitrary number of observed contexts, referred to as the context dataset $\c{D} = \{(x_i, y_i)\}_{i=1}^n$. 

NPs use an encoder-decoder architecture to define the conditional distributions. The encoder is a NN which operates on the context dataset to output a dataset representation $r=enc(\c{D})$. Using this representation, a decoder predicts the function output at a test location. Conditional NPs \citep{garnelo2018conditional} define the target distribution as a Gaussian which factorises over the different target points 
\begin{equation}
\label{eq:app:cnp}
    p(\v{y}^* \given \v{x}^*, \c{D}) = \prod_{i=1}^n \c{N}(y_i^* \given dec_{\mu}(x_i^*, r), dec_{\sigma^2}(x_i^*, r)),
\end{equation}
where the mean and variance of the Gaussians are given by decoding the context dataset representation $r$ and targets $x_i^*$. When the context dataset is empty $\c{D} = \varnothing$, NPs typically set the representation $r$ to a fixed vector.

A latent NP~\citep{garnelo2018neural} relies on a similar encoder-decoder architecture but now the encoder is used to parameterise a global latent variable $z$. Conditioned on $z$ the likelihood factorises over the target points
\begin{equation}
\label{eq:app:lnp}
    p(\v{y}^* \given \v{x}^*, \c{D}) = \int_z \prod_{i=1}^n \c{N}\left(y_i^* \given dec_{\mu}(x_i^*, r, z), dec_{\sigma^2}(x_i^*, r, z)\right) p(z \given r) \d z.
\end{equation}
The idea behind having a global latent variable is to model various instances of the stochastic process.

The encoder and decoder networks are trained by maximising a lower bound to the log likelihood over different function realisations, where a random subset of points is placed in the context and target sets. In all experiments we use an existing opensource package \url{https://github.com/wesselb/neuralprocesses} for the NP baselines as it provides well-tested and finely-tuned model configurations for a variety of tasks.

\section{On Marginal Consistency of Neural Processes and Neural Diffusion Processes}
\label{sec:app:consistency}

When discussing marginal consistency, there are two different settings that one needs to consider.
\begin{enumerate}
    \item For a given context dataset, does marginal consistency,~$p(y_0 \given \c{D}) = \int p(y_0, y_1,\ldots,y_n \given \c{D}) \d y_{1:n}$, hold?
    \item Are the conditional distributions related through Bayes' rule: $p(y_0 \given \c{D}) = \int p(y_0 \given \c{D}) p(y_1 \given y_0, \c{D})\d y_1 $?
\end{enumerate}

It is straightforward to prove that the family of NP models satisfy the first condition. This follows directly from the factorisation of the target distribution in \cref{eq:app:cnp,eq:app:lnp}. We refer to \citet{dubois2020npf} for a concise overview of the proofs. As a result, NPs will for a given context dataset satisfy the KET conditions. However, it is important to note that consistency of NPs is not satisfied within contexts \citep{kim2019attentive}. To see this, consider the case where we marginalise over $y_1$ and apply Bayes' rule
\begin{equation}
\label{eq:context-consistency-nps}
    p(y_0 \given \c{D}) = \int_{y_1} p(y_1 \given \c{D})\, p(y_0 \given y_1, \c{D}) \d y_1.
\end{equation}
The distribution produced by the NP on the LHS of the equation need not match the distribution one would obtain if you append the additional point $(x_1, y_1)$ to the context dataset on the RHS. This is a consequence of the NNs, which are responsible for encoding and decoding the context dataset, and can not guarantee consistency when their input changes. This holds true for latent and conditional NPs, which means neither of them are marginally consistent when their context dataset changes.

NDPs can not mathematically guarantee marginal consistency as defined by \cref{eq:marginal-consistency} for an arbitrary number of points $n$ and sequences $\{x_i\}_{i=1}^n$. Hence, they do not satisfy the criteria for Kolmogorov Extension Theorem. However, it is important to note that NDPs differ from NPs in their approach to constructing a predictive distribution. In NPs, the predictive distribution is generated directly by the NN. In NDPs, on the contrary, the predictive distribution is constructed through conditioning the joint.

\subsection{Empirical Evaluation}

\begin{table}[!tb]
  \centering
  \caption{Variability in the violin plots of~\cref{fig:consistency-violin} measured by the standard deviation on the means.\label{tab:consistency}}
\begin{tabular}{lrrrrr}
\toprule
model & $\mu$ & $\sigma^2$ & $q_{50}$ & $q_{90} - q_{10}$ & $q_{75} - q_{25}$  \\
\midrule
GP    &  0.005687 &  0.001491 &  0.005245 &      0.010979 &      0.011858 \\
NDP   &  0.019796 &  0.002985 &  0.018961 &      0.018718 &      0.010045 \\
\bottomrule
\end{tabular}
\end{table}

While NDPs do not mathematically satisfy marginal consistency for arbitrary sequences, the following experiment demonstrates that they approximate consistency through meta-learning. Similar to NPs, we argue that this behaviour is induced by the maximum log-likelihood objective, which can be seen as minimising the KL between the consistent samples from the data-generating process and the NDP's output.

\begin{figure}[!tb]
\centering
\includegraphics[width=\linewidth]{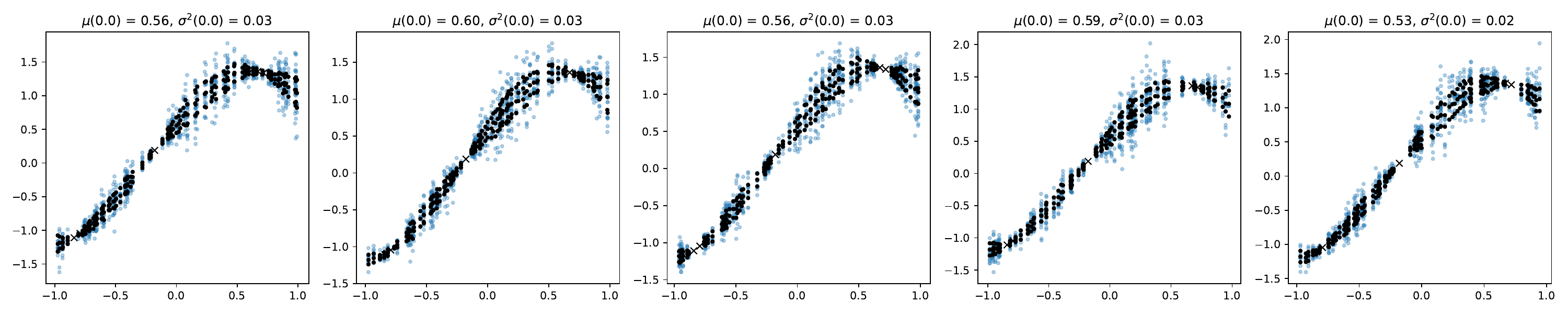}
\caption{Samples (blue) from the NDP with empirical 25 and 75th quantiles (black). In each of the 5 plots we vary the locations at which the posterior is evaluated. We make sure that $x=0$ is included in the test points, which allows us to compute the empirical mean, variance and quantiles of $y_0$, which corresponds to the sample evaluated at $x=0$.} %
\label{fig:consistency-samples}
\end{figure}

To illustrate this behaviour, let us sample from the joint $p(y_0 \cup \{y_i\}_{i=1}^{n}) \given \c{D})$, and investigate the dependence of the marginal $p(y_0)$ as we vary the other test inputs $\{x_i\}_{i=1}^{n}$. We set $y_0$ to correspond to the prediction at $x=0$ and study 5 different random configurations for the other input locations $\{x_i\}_{i=1}^{n}$, numbered \#1 to \#5. In \cref{fig:consistency-samples}, we show the posterior samples $p(y_0 \cup \{y_i\}_{i=1}^{n} \given \c{D})$, the difference between the 5 plots is that we evaluate the posterior at different locations $\{x_i\}_{i=1}^{n}$ in $[-1, 1]$. For each configuration we make sure that $x=0.0$ is included such that we can investigate the marginal $p(y_0)$.

\begin{figure}[!tb]
\centering
\includegraphics[width=\linewidth]{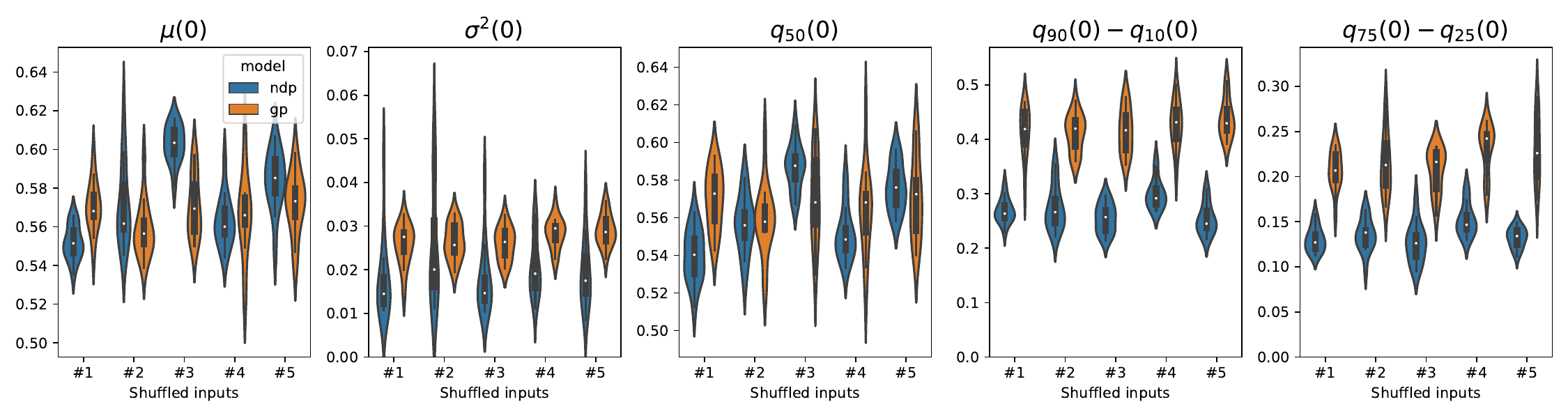}
\caption{Boxplot of the empirical mean, variance, median and quantile widths of \cref{fig:consistency-samples} by repeating the experiment 20 times for each random input configuration. We compare the NDP across different input configurations and agains a GP.}
\label{fig:consistency-violin}
\end{figure}

\Cref{fig:consistency-violin} shows the empirical distribution across samples of the mean $\mu$, variance $\sigma^2$, median $q_{50}$ and quantile widths of $p(y_1)$ for each of the different input configurations. As a reference, we also add the statistics of GP at $x=0$ samples, which are drawn from a consistent distribution, to the figure. Consistency is observed when the violin plots of the statistics do not change as of a result of varying $\{x\}_{i=1}^n$ (i.e., \#1, \#2, ..., \#5 for each model appear the same). 

We can quantify this variability by computing the standard deviation on the means of the violin plots across the varying inputs. The numerical values of this are given in \cref{tab:consistency}. We observe that variation in the quantiles are found to be lower than 10\%. By comparing with the equivalent performance of a GP, we find that the bulk of the variations are due to stochastic fluctuations as a result of finite sample size as opposed to the inconsistencies associated with the NDP.

\section{Additional Information on the Experiments}
\label{sec:app:experiments}

\subsection{Experimental Setup and Overview}
\label{sec:app:experimental-setup}

Here we provide more detailed information describing how the numerical experiments were conducted.

All experiments share the same model architecture illustrated in \Cref{fig:architecture}, there are however a number of model parameters that must be chosen. An L1 (i.e.,~Mean Absolute Error, MAE) loss function was used throughout. We use four or five bi-dimensional attention blocks, each consisting of multi-head self-attention blocks \citep{vaswani2017attention} containing a representation dimensionality of $H=64$ and $8$ heads. Each experiment used either $500$ or $1000$ diffusion steps, where we find  larger values produce more accurate samples at the expense of computation time. Following \citet{nichol2021glide} we use a cosine-based scheduling of $\beta_t$ during training. The Adam optimiser is used throughout. Our learning rate follows a cosine-decay function, with a $20$ epochs linear learning rate warm-up to a maximum learning rate of $\eta = 0.001$ before decaying. 
All NDP models were trained for $250$ epochs with the exception of the lengthscale marginalisation experiment, which was trained for $500$ epochs. Each epoch contained $4096$ example training $(\v{y}_0, \v{x}_0)$ pairs. Training data was provided in batches of $32$, with each batch containing data with the same kernel hyperparameters but different realisations of prior GP samples. The complete configuration for each experiment is given in \cref{table:experiment-configuration}

Experiments were conducted on a 32-core machine and utilised a single Tesla V100-PCIE-32GB GPU. Training of each model used in the experiments takes no longer than 30 minutes, except for the image regression experiments.

\begin{table}[bth]
  \caption{Experiment configuration and training time. \label{table:experiment-configuration}}
\centering
\begin{tabular}{lcccccc} \toprule
    {Experiment} & {Symthetic data} & {Hyperparameter} & {Step} & {High dim} & {1D opt} & {Image} \\
    regression & & marginalisation & & BO & & regression \\ \midrule
    Epochs & 250 & {500} & 250  & 250  & 250 & 100 \\
    Total samples seen & 1024k & {2048k} & 1024k & 1024k & 1024k & - \\ 
    Batch size & 32 & 32 & 32 & 32 & 32  & 32 \\
    Loss & L1 & L1 & L1 & L1 & L1  & L1 \\
    LR decay  & cosine  & cosine & cosine  & cosine & cosine & cosine \\
    LR init  & $2e^{-5}$  & $0.001$ & $0.001$ & $0.001$ & $0.001$ & $2e^{-5}$\\
    LR warmup epochs & 20 & 20 & 20 & 20 & 20  & 20 \\
    Num blocks & 4  & 5 & 5 & 5 & 5 & 5 \\
    Representation dim ($H$) & 64 & 64 & 64 & 64 & 64 & 64 \\
    Num heads & 8 & 8 & 8 & 8 & 8 & 8 \\
    Num timesteps (T) & 500 & {1000} & {1000} & 500 & {2000} & 500 \\
    $\beta$ schedule  & cosine  & cosine & cosine  & cosine & cosine & cosine \\
    Num points (N) & 60/70/80 & 100 & 100 & {256} & 100 & $H\times W$\\
    Deterministic inputs & True & True & True & True & {False} & True \\ \midrule
    Training time & 17m & 33m & 16m & 21m & 16m & 10h \\ \bottomrule
\end{tabular}
\end{table}

\paragraph{Time embedding}
\label{sec:app:time}
The diffusion step $t$ is a crucial input of the NN noise estimator as the model needs to be able to differentiate between noise added at the start or the end of the process. Following \citet{vaswani2017attention} we use a cyclic 128-dimensional encoding vector for each step 
\begin{equation*}
t \mapsto
        [\sin(10^{\frac{0 \times 4}{63}} t),
        \sin(10^{\frac{1 \times 4}{63}} t),
        \ldots,
        \sin(10^{\frac{64 \times 4}{63}} t),
        \cos(10^{\frac{0 \times 4}{63}} t),
        \ldots,
        \cos(10^{\frac{64 \times 4}{63}} t)] \in \R^{128}
\end{equation*}

\subsection{Illustrative Figure}

\begin{figure}[t]
\centering
\begin{subfigure}{\textwidth}
\centering
\includegraphics[width=\textwidth]{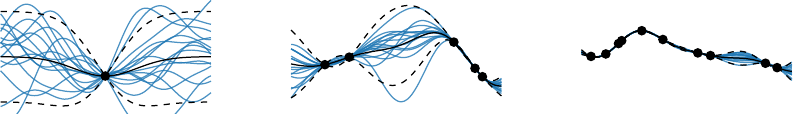}
\caption{GP Regression}
\label{fig:one-d-regression-gp}
\end{subfigure}
\begin{subfigure}{\textwidth}
\centering
\includegraphics[width=\textwidth]{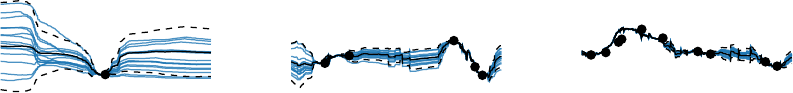}
\caption{Attentive Latent Neural Process}
\label{fig:one-d-regression-np}
\end{subfigure}
\begin{subfigure}{\textwidth}
\centering
\includegraphics[width=\textwidth]{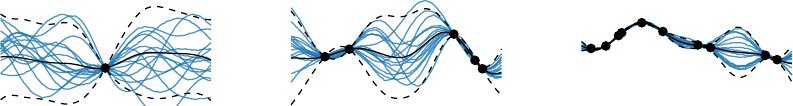}
\caption{Neural Diffusion Process (ours)}
\label{fig:one-d-regression-ndp}
\end{subfigure}
\caption{\textbf{1D regression:} The blue curves are posterior samples from different probabilistic models. We also plot the empirical mean and two standard deviations of the samples in black. From left to right we increase the number of data points (black dots) and notice how the process gets closer to the true underlying function.}
\label{fig:one-d-regression}
\end{figure}

For the creation of \cref{fig:one-d-regression-single,fig:one-d-regression}, we use GPflow \citep{gpflow2020} for the GP regression model using a kernel that matches the training data: a squared exponential with lengthscale set to $0.2$. The other baseline, the Attentive Latent NP, is a pre-trained model which was trained on a dataset with the same configuration. The ALNP model originates from the reference implementation of \citet{dubois2020npf}.

\subsection{Hyperparameter Marginalisation}

In this experiment, we provide training data to the NDP in the form of unique prior samples from a ground truth GP model. The input for each prior sample, $\v{x}_0$, is deterministically spaced across $[-1,1]$ with $N=100$. The corresponding output $\v{y}_0$ is sampled from a GP prior $\v{y}_0 | \v{x}_0 \sim \c{N}\big(\v{0}, k(\v{x}_0, \v{x}_0) + \sigma^2 \big)$ where the kernel is a stationary Matern-$\frac{3}{2}$ kernel. The noise variance is set to $\sigma^2 = 10^{-6}$ and the kernel variance is fixed to $\sigma_k^2=1.0$ throughout. We place a log-normal prior on the lengthscale $\log \c{N}(\log 0.5, \sqrt{0.5})$.

\subsection{Non-Gaussian Posteriors}

The step function training data is created synthetically by drawing $u \~ \c{U}[-1,1]$ and letting
\begin{equation}
    f(x) = 0\text{ for } x \le u, \qquad\text{and}\qquad f(x) =1\text{ for } x>u.
\end{equation}

\begin{figure}[t]
\centering
\begin{subfigure}{\textwidth}
\centering
\includegraphics[clip,trim=10cm 0cm 10cm 0cm,width=.8\textwidth]{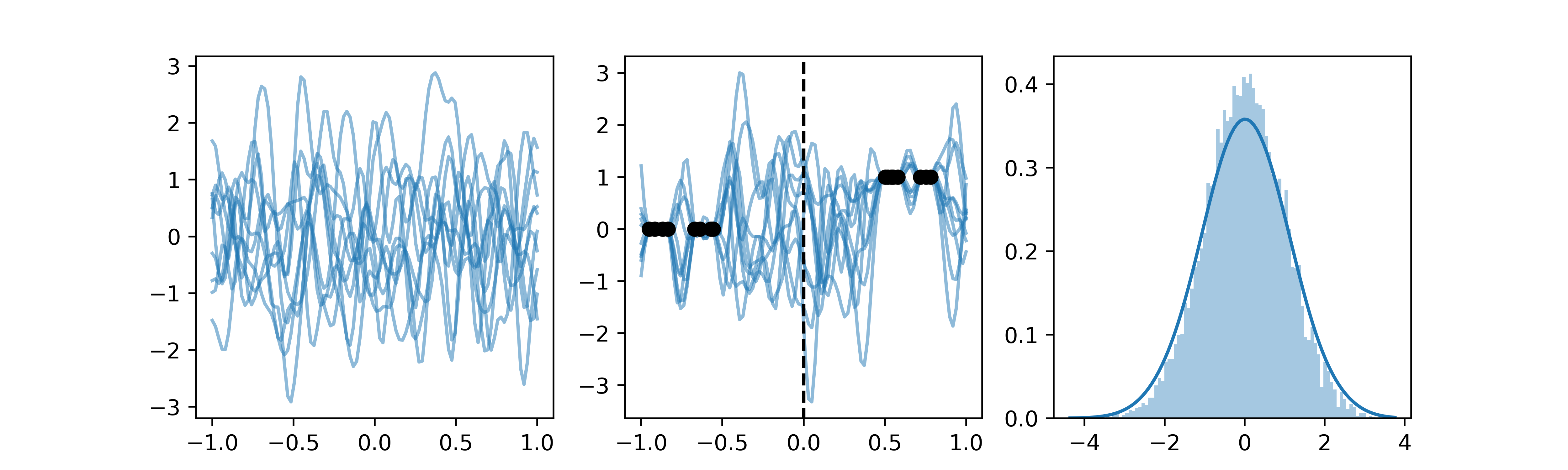}
\caption{Squared Exponential kernel}
\end{subfigure}
\begin{subfigure}{\textwidth}
\centering
\includegraphics[clip,trim=10cm 0cm 10cm 0cm,width=.8\textwidth]{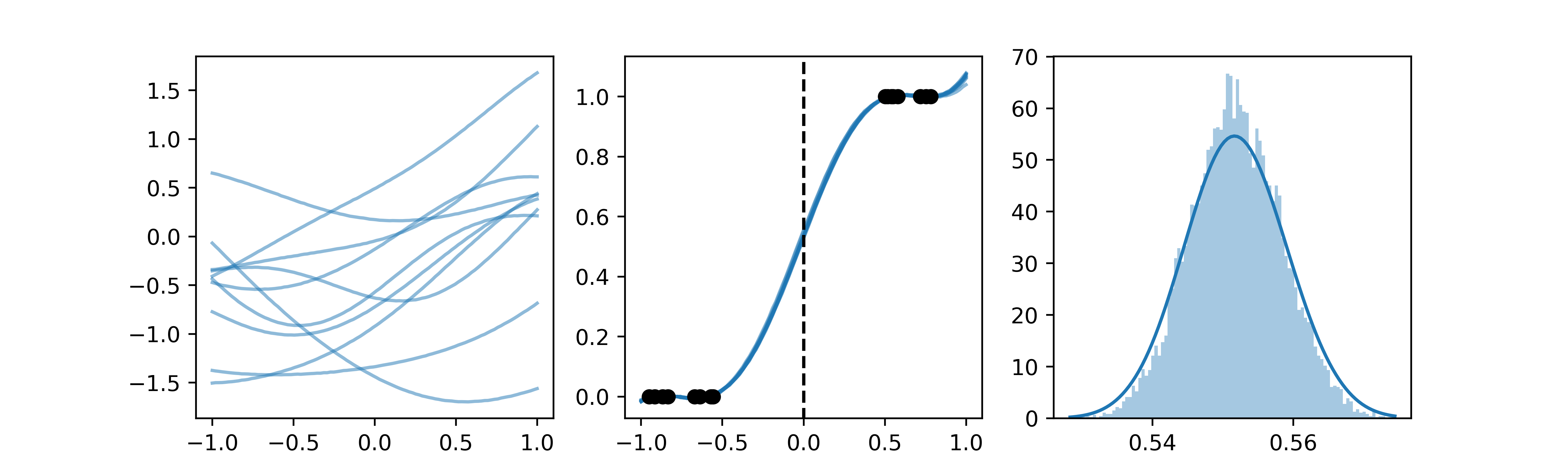}
\caption{Spectral Mixture kernel}
\end{subfigure}
\caption{Performance of Gaussian process models on step function data. The left panel shows samples from the prior. The middle panel shows conditional samples where the data is given by the black dots. The right panel shows the marginal at $x=0$.}
\label{fig:step-function-gps}
\end{figure}

For completeness, we show the performance of two GP models on this data in \cref{fig:step-function-gps}. As per the definition, each marginal of a GP is Gaussian which makes modelling this data impossible.

\subsection{Image Regression}
\label{sec:app:image-regression}

In this experiment we train a NDP on two image datasets: MNIST and CelebA. We treat an image as a function with 2D inputs in $[1, H] \times [1, W]$ and 1D grey-scale outputs (MNIST) or 3D RGB output (CelebA). For both datasets we re-scale the 2D inputs to the unit box $[-2, 2] \times [-2, 2]$ and the outputs to normalised. %

\subsection{Regression on Synthetic Data}

\begin{figure}
\centering
\begin{subfigure}{\textwidth}
\centering
\includegraphics{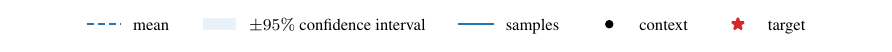}
\end{subfigure}
\begin{subfigure}{\textwidth}
\includegraphics[width=\linewidth]{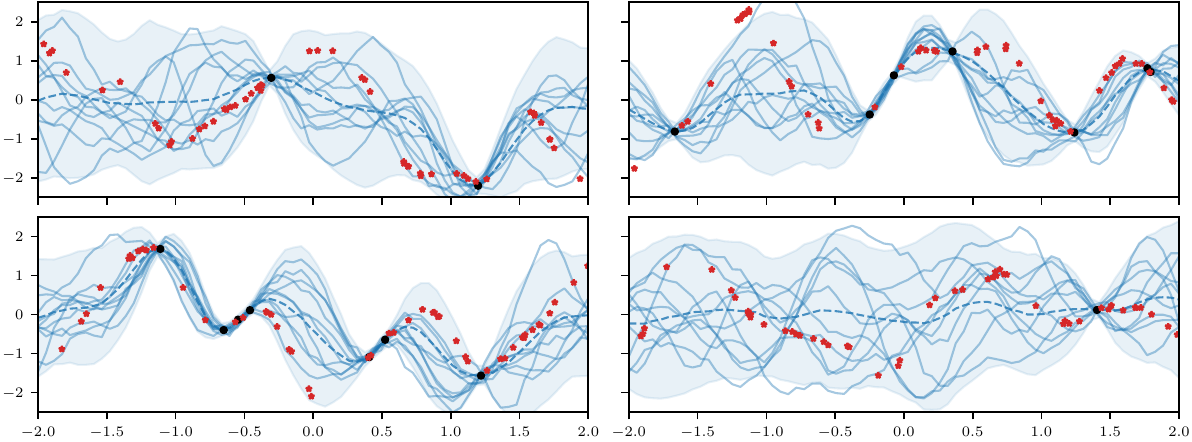}
\caption{Squared Exponential}
\vspace{5mm}
\end{subfigure}
\begin{subfigure}{\textwidth}
\includegraphics[width=\linewidth]{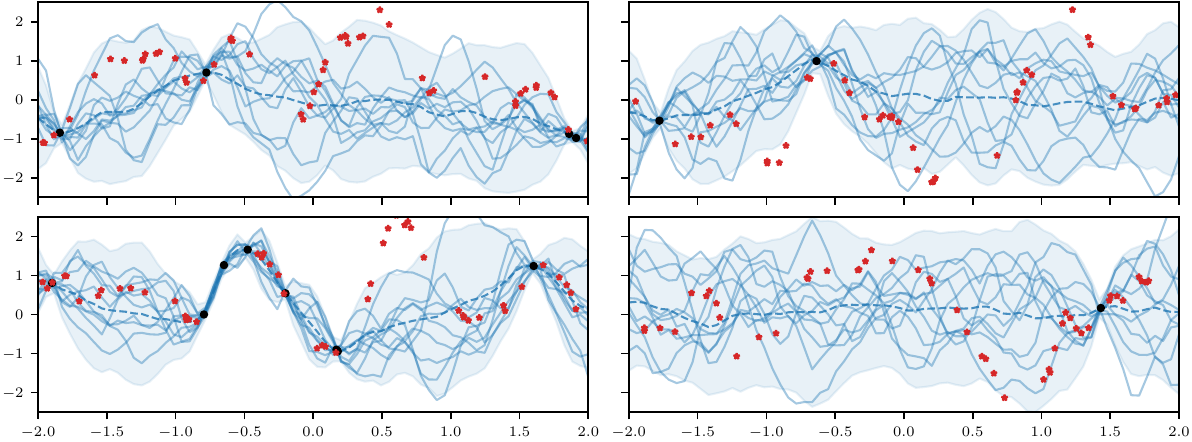}
\caption{Mat\'ern--$\frac52$}
\end{subfigure}
\caption{Model predictions on 1D synthetic datasets. Black dots: random number of context points. Red crosses: 50 target points. Blue lines: samples of the model, mean and 95\% confidence intervals using the samples. \label{fig:app:synthetic-regression}}
\end{figure}

We adopt the same experimental setup as \citet{wang2012gaussian} to generate synthetic data, which includes Gaussian process (Squared Exponential ({\scshape SE}), {\scshape Mat'ern-$\frac52$}) sample paths. \Cref{fig:app:synthetic-regression} displays samples from each of these datasets, which are corrupted with observation noise having a variance of $\sigma^2=0.05^2$. We use a single lengthscale across all dimensions set to $\ell = 0.25 \sqrt{D}$.

The training data is composed of $2^{14}$ sample paths, whereas the test dataset comprises $128$ paths. For each test path, we sample the number of context points within a range of $1$ and $10 D$. We fix the number of target points to $50$. The input range for both the training and test datasets, covering both context and target sets, is set to $[-2, 2]$.

\subsection{Bayesian Optimisation}
\label{sec:app:high-dim-bo}

In this experiment, we perform Bayesian optimisation on the Hartmann 3D \& 6D, Rastrigin 4D and Ackley 5D objectives. We re-scale, without loss of generality, the inputs of the objectives such that the search space is $[-1, 1]^D$. %

The baseline models, GPR and Random, originate from Trieste \citep{berkeley2022trieste} ---a TensorFlow/GPflow-based Bayesian Optimisation Python package. We benchmark two NDP models: Fixed and Marginalised. The Fixed NDP model is trained on Mat\'ern-$\frac52$ samples with a fixed lengthscale set to $0.5$ along all dimensions. The Marginalised NDP model is trained on Mat\'ern-$\frac52$ samples originating from different lengthscales, drawn from a log-Normal prior.

\end{document}